%% file: main.tex
\documentclass[a4paper]{cas-sc}

\usepackage[all]{foreign}
\usepackage[numbers]{natbib}
\usepackage[capitalize]{cleveref}
\usepackage{booktabs}
\usepackage{multirow}
\usepackage{url}
\usepackage{bm}
\usepackage{siunitx}
\usepackage{subcaption}
\usepackage{glossaries}
\usepackage{tabularx}
\usepackage{xspace}

\glsdisablehyper
\input{acronyms}
\input{commands}

\begin{document}

\let\WriteBookmarks\relax
\def\floatpagepagefraction{1}
\def\textpagefraction{.001}

\shorttitle{From Activity Recognition to Shift Detection: Variational Wi-Fi Sensing with Evidential Uncertainty}
\shortauthors{Cotti, Cominelli et~al.}

\title[mode = title]{From Activity Recognition to Shift Detection: Variational Wi-Fi Sensing with Evidential Uncertainty}
\tnotemark[1]
\tnotetext[1]{This work was supported by the European Office of Aerospace Research \& Development (EOARD) under award number \#FA8655\--22\--1\--7017; the U.S. Army DEVCOM Army Research Laboratory (ARL) under Cooperative Agreements \#W911NF2220243 and \#W911NF1720196. Any opinions, findings, and conclusions or recommendations expressed in this material are those of the authors and do not necessarily reflect the views of the United States government.}

\author[1]{Luca Cotti}[orcid=0009--0004--6351--556X]
\ead{luca.cotti@unibs.it}
\credit{Methodology, Software, Validation, Formal Analysis, Investigation, Data Curation, Writing \-- Original Draft, Writing \-- Review \& Editing, Visualization}

\author[2]{Marco Cominelli}[orcid=0000--0002--1838--348X]
\ead{marco.cominelli@polimi.it}
\credit{Conceptualization, Methodology, Software, Validation, Formal Analysis, Investigation, Resources, Data Curation, Writing \-- Original Draft, Writing \-- Review \& Editing, Visualization}

\author[1]{Francesco Gringoli}[orcid=0000--0003--2621--582X]
\ead{francesco.gringoli@unibs.it}
\credit{Resources, Writing \-- Review \& Editing}

\author[3]{Lance M. Kaplan}[orcid=0000--0002--3627--4471]
\ead{lance.m.kaplan.civ@army.mil}
\credit{Conceptualization, Validation, Writing \-- Review \& Editing}

\author[4]{Mani B. Srivastava}[orcid=0000--0002--3782--9192]
\ead{mbs@ucla.edu}
\credit{Conceptualization, Methodology, Validation, Writing \-- Review \& Editing}

\author[5]{Trevor Bihl}[orcid=0000--0003--2431--2749]
\ead{tbihl@shawnee.edu}
\credit{Validation, Writing \-- Review \& Editing}

\author[6]{Erik Blasch}[orcid=0000--0001--6894--6108]
\ead{erik.blasch.1@us.af.mil}
\credit{Validation, Writing \-- Review \& Editing, Project Administration, Funding Acquisition}

\author[7]{Kara Combs}[orcid=0000--0002--6313--8563]
\ead{kara.combs.1@us.af.mil}
\credit{Validation, Writing \-- Review \& Editing, Project Administration, Funding Acquisition}

\author[6]{Daniel O. Brigham}[orcid=0000--0002--3630--8483]
\ead{daniel.brigham@us.af.mil}
\credit{Validation, Writing \-- Review \& Editing, Project Administration, Funding Acquisition}

\author[1,8,9,10]{Federico Cerutti}[orcid=0000--0003--0755--0358]
\cormark[1]
\ead{federico.cerutti@unibs.it}
\credit{Conceptualization, Methodology, Validation, Resources, Writing \-- Review \& Editing, Supervision, Project Administration, Funding Acquisition}

\cortext[1]{Corresponding author}

\affiliation[1]{organization={Department of Information Engineering, University of Brescia},
  addressline={Via Branze 38},
  city={Brescia},
  postcode={25123},
  state={Lombardy},
  country={Italy}}

\affiliation[2]{organization={Department of Electronics, Information and Bioengineering, Politecnico di Milano},
  addressline={Piazza Leonardo da Vinci 32},
  city={Milan},
  postcode={20133},
  state={Lombardy},
  country={Italy}}

\affiliation[3]{organization={U.S. Army DEVCOM Army Research Laboratory},
  addressline={2800 Powder Mill Road},
  city={Adelphi},
  postcode={20783},
  state={Maryland},
  country={USA}}

\affiliation[4]{organization={Electrical and Computer Engineering Department, University of California},
  addressline={405 Hilgard Ave},
  city={Los Angeles},
  postcode={90095},
  state={California},
  country={USA}}

\affiliation[5]{organization={College of Business and Engineering, Shawnee State University},
  addressline={940 2nd St},
  city={Portsmouth},
  postcode={45662},
  state={Ohio},
  country={USA}}
  
\affiliation[6]{organization={Air Force Research Laboratory},
  addressline={1864 4th Street},
  city={Wright-Patterson AFB},
  postcode={45433},
  state={Ohio},
  country={USA}}

\affiliation[7]{organization={Air Force Research Laboratory},
  addressline={2241 Avionics Circle},
  city={Wright-Patterson AFB},
  postcode={45433},
  state={Ohio},
  country={USA}}

\affiliation[8]{organization={Cardiff University},
  addressline={Cardiff CF10 3AT},
  city={Cardiff},
  country={UK}}

\affiliation[9]{organization={University of Southampton},
  addressline={University Road},
  city={Southampton},
  postcode={SO17 1BJ},
  country={UK}}

\affiliation[10]{organization={Imperial College London},
  addressline={Exhibition Rd},
  city={London},
  postcode={SW7 2AZ},
  country={UK}}

\glsresetall

\begin{abstract}
  Human motion perturbs wireless channels, making commodity Wi-Fi devices a practical platform for device-free activity recognition. Yet Channel State Information (CSI) also depends strongly on the person, receiver position, recording session, and surrounding environment, so a classifier can continue producing activity labels after its deployment conditions have moved away from those seen during training. We propose Wi-HAR, a multi-antenna architecture in which four Variational Auto-Encoders learn compact distributions from raw Wi-Fi 6 CSI magnitudes and a delayed-fusion classifier combines their means and log-variances. On twelve activities from a public benchmark, Wi-HAR reaches 96.6\% median recognition accuracy, compared with 70.8–81.5\% for independently optimized single-antenna models, showing that the antennas provide complementary views of the same activity. We then use the same latent backbone to detect deployment shifts without using shifted samples for model selection or calibration. A class-conditional uncertainty score derived from Evidential Deep Learning achieves a macro-average \gls{roc}\--\gls{auc} of 0.897 across eleven datasets that change receiver position, participant, recording session, or room, and detects 82.5\% of shifted samples at a realized 5.2\% in-distribution rejection rate. Detection performance nevertheless varies substantially across receiver positions, showing that the same physical change does not induce a uniform uncertainty response. The compact representation also supports shallow decision-tree analysis using variables from several antennas. 
\end{abstract}


\maketitle

\glsresetall
\section{Introduction}\label{sec:introduction}
\input{parts/1_introduction}

\section{Background}\label{sec:background}
\input{parts/2_background}

\section{Methodology and Hypotheses}\label{sec:methodology}
\input{parts/3_methodology}

\section{Activity Recognition Results}\label{sec:results}
\input{parts/4_results}

\section{Uncertainty-Aware Detection of Deployment Shifts}\label{sec:uncertainty}
\input{parts/5_uncertainty}

\section{Conclusion and Future Challenges}\label{sec:conclusions}
\input{parts/6_conclusions}

\section*{Declaration of Generative AI Use}
During preparation of this work, the authors used GPT 5.6 to revise and improve the fluency of the text. The authors subsequently reviewed and edited the content and take full responsibility for the article.

\section*{Declaration of Competing Interest}
The authors declare that they have no known competing financial interests or personal relationships that could have appeared to influence the work reported in this paper.

\printcredits

\bibliographystyle{cas-model2-names}
\bibliography{references}

\end{document}

%% file: acronyms.tex
\newacronym{cnn}{CNN}{Convolutional Neural Network}
\newacronym{csi}{CSI}{Channel State Information}
\newacronym{edl}{EDL}{Evidential Deep Learning}
\newacronym{gen}{GEN}{Generative Evidential Neural Network}
\newacronym{har}{HAR}{Human Activity Recognition}
\newacronym{idft}{IDFT}{Inverse Discrete Fourier Transform}
\newacronym{kl}{KL}{Kullback-Leibler}
\newacronym{mlp}{MLP}{Multi-Layer Perceptron}
\newacronym{ood}{OoD}{out-of-distribution}
\newacronym{ofdm}{OFDM}{Orthogonal Frequency-Division Multiplexing}
\newacronym{vae}{VAE}{Variational Auto-Encoder}
\newacronym{elbo}{ELBO}{Evidence Lower Bound}
\newacronym{gelu}{GELU}{Gaussian Error Linear Unit}
\newacronym{tpe}{TPE}{Tree-structured Parzen Estimator}
\newacronym{tsne}{t-SNE}{t-Distributed Stochastic Neighbor Embedding}
\newacronym{gan}{GAN}{Generative Adversarial Network}
\newacronym{roc}{ROC}{Receiver Operating Characteristic}
\newacronym{cdf}{CDF}{Cumulative Distribution Function}
\newacronym{pdf}{PDF}{Probability Density Function}
\newacronym{auc}{AUC}{Area Under the Curve}
\newacronym{id}{ID}{In-Distribution}
\newacronym{rope}{ROPE}{Region of Practical Equivalence}

%% file: commands.tex
\newcommand{\name}{Wi-HAR\xspace}
\newcommand{\wifi}{\mbox{Wi-Fi}\xspace}

\newcommand{\bb}[1]{\mathbf{#1}}
\newcommand{\bx}{\bb{x}}
\newcommand{\bxi}{\bx^{(i)}}

\newcommand{\bz}{\bb{z}}
\newcommand{\bzi}{\bz^{(i)}}
\newcommand{\bzil}{\bz^{(i,l)}}
\newcommand{\bT}{\boldsymbol{\theta}}
\newcommand{\bI}{\bb{I}}
\newcommand{\bX}{\bb{X}}
\newcommand{\bphi}{\boldsymbol{\phi}}
\newcommand{\beps}{\boldsymbol{\epsilon}}
\newcommand{\bepsl}{\beps^{(l)}}
\newcommand{\bsigma}{\boldsymbol{\sigma}}
\newcommand{\bmu}{\boldsymbol{\mu}}
\newcommand{\bzero}{\bb{0}}
\newcommand{\pT}{p_{\bT}}

\newcommand{\Exp}[2]{\mathbb{E}_{#1}\left[#2\right]}
\newcommand{\LB}[2]{\mathcal{L}^{#1}(\bT,\bphi; #2)}
\newcommand{\tsignal}{\ensuremath{x}\xspace}
\newcommand{\tsymbol}{\ensuremath{x_n}\xspace}
\newcommand{\fsymbol}{\ensuremath{\mathbf{x}_n}\xspace}
\newcommand{\fsymbolrec}{\ensuremath{\hat{\mathbf{x}}_n}\xspace}
\newcommand{\numsymbols}{\ensuremath{N}\xspace}
\newcommand{\numsubcarr}{\ensuremath{M}\xspace}
\newcommand{\nth}{$n$\textsuperscript{th}\xspace}
\newcommand{\channel}{\ensuremath{\mathcal{H}}\xspace}
\newcommand{\csi}{\ensuremath{\mathbf{H}}\xspace}
\newcommand{\yfsymbol}{\ensuremath{\mathbf{y}_n}\xspace}
\newcommand{\yfsymbolant}{\ensuremath{\mathbf{y}^{(l)}_{n}}\xspace}
\newcommand{\complexm}{\mathbb{C}^{\numsubcarr \times 1}}
\DeclareMathOperator{\diag}{diag}

%% file: parts/1_introduction.tex
Human motion changes the multipath propagation of \wifi signals and therefore leaves measurable signatures in the \gls{csi} observed by a receiver. These signatures make commodity \wifi devices a practical platform for passive and device-free sensing, including gesture recognition, motion tracking, and human activity recognition~\cite{liPassiveWiFiRadar2021}. Recent \wifi~6 (\ie, IEEE 802.11ax) devices expose particularly fine-grained \gls{csi} measurements~\cite{storrerIndoorTrackingMultiple2021}, further increasing the amount of information available for sensing.

The same dependence on the propagation environment that makes \gls{csi} useful for sensing also makes recognition models sensitive to deployment conditions. Changes in receiver position, participant, recording session, or room geometry alter the multipath channel and can move the observed \gls{csi} away from the distribution seen during training~\cite{cominelliExposingCSISystematic2023}. A classifier may therefore continue producing activity labels even after the physical conditions underlying those predictions have changed. High recognition accuracy under the original collection condition is insufficient for deployment: a sensing system should recognize activities under familiar conditions and indicate when its inputs no longer resemble those conditions.

We address these two requirements with \name, a multi-antenna architecture based on \glspl{vae}~\cite{kingmaAutoencodingVariationalBayes2022}. Four antenna-specific encoders independently map raw \gls{csi} magnitudes to compact Gaussian latent representations, and a delayed-fusion classifier combines their means and log-variances for activity recognition. The use of multiple antennas is central to the architecture because physically separated antennas observe different multipath channels and can therefore provide complementary views of the same activity. At the same time, the representation is deliberately compact: unlike autoencoder-based \gls{har} systems that use latent spaces with tens of neurons~\cite{seyfiogluDeepConvolutionalAutoencoder2018}, \name represents each antenna with only 2 to 12 distribution parameters. This compression keeps the fused representation small enough to examine directly while retaining sufficient information for activity discrimination.

We use the same latent backbone to determine whether the deployment conditions have shifted. We compare the conventional Softmax classifier with \gls{edl}~\cite{sensoyEvidentialDeepLearning2018} and \gls{gen}~\cite{sensoyUncertaintyawareDeepClassifiers2020}, and construct a class-conditional anomaly score that measures how unusual a prediction's uncertainty is relative to training samples assigned to the same activity. We evaluate this score across eleven datasets that change collector position, participant, recording session, or room. No shifted dataset contributes to model selection, construction of the uncertainty reference, or threshold calibration, so the experiment tests whether uncertainty can expose previously unseen changes in sensing conditions rather than learn the particular shifts contained in the benchmark.

The experiments support three main findings. First, the compact multi-antenna representation reaches a median activity-recognition accuracy of 96.6\%, whereas independently optimized single-antenna models reach 70.8--81.5\%, showing that the four channels provide complementary information about the observed activities. Second, uncertainty in the same latent backbone provides a useful signal of changed deployment conditions: \gls{edl} attains a macro-average \gls{roc}\--\gls{auc} of 0.897 across the eleven shifted datasets and detects 82.5\% of shifted samples at a realized 5.2\% in-distribution rejection rate. Detection performance nevertheless varies substantially across collector positions, showing that the same category of physical change does not produce a uniform uncertainty response. Third, the compact representation remains amenable to direct analysis: a depth-four decision tree reaches 96.5\% accuracy on six activities and bases its decisions on latent variables from several antennas, providing an interpretable counterpart to the multi-antenna ablation results.

This article substantially extends the conference study in~\cite{cominelliAccuratePassiveRadar2023}. We expand the recognition task to all twelve activities in the public benchmark, search a broader set of encoder and fusion architectures, and compare multi-antenna fusion with independently optimized single-antenna models. More importantly, we extend \name from closed-set activity recognition to deployment-shift detection by introducing the class-conditional uncertainty analysis described above, and evaluating both threshold-free and conformally calibrated operating points across eleven controlled changes in sensing conditions. We also examine the learned representation with a shallow decision tree. Together, these experiments test whether a compact variational representation can support accurate activity recognition while also providing an explicit signal that the deployment conditions underlying the classifier have changed.

%% file: parts/2_background.tex
\subsection{CSI Data Measurement}\label{sec:csi-details}

Let \tsignal denote the signal transmitted by a generic \wifi device. Within each \wifi frame, \tsignal contains \numsymbols consecutive \gls{ofdm} symbols \tsymbol ($n=1,\dots,\numsymbols$), each spanning \numsubcarr orthogonal narrow-band subcarriers. To construct the \nth symbol in the frequency domain, the transmitter maps the input bits onto a vector of \numsubcarr complex modulation symbols $\fsymbol \in \complexm$ and then applies the \gls{idft} to obtain the time-domain waveform \tsymbol.

The value of \numsubcarr depends on both the transmission bandwidth and the \wifi standard. For example, \wifi~4 (IEEE~802.11n) allocates $M=64$ subcarriers over \SI{20}{\mega\hertz} channels, whereas \wifi~6 (IEEE~802.11ax) can allocate up to $M=2048$ subcarriers over \SI{160}{\mega\hertz} channels.

Wireless channels are generally frequency-selective and therefore affect each subcarrier $\fsymbol[m]$ differently. Neglecting noise and interference, we express the \nth received \gls{ofdm} symbol in the frequency domain as
\begin{equation}
    \yfsymbol = \channel \odot \fsymbol \, ,
    \label{eq:freq-channel}
\end{equation}
where $\channel \in \complexm$ is the frequency response of the wireless channel and $\odot$ is the element-wise product (\ie Hadamard product) that expresses the channel effect on the $M$ subcarriers of the \gls{ofdm} frame.

To equalize the received signal, the receiver must estimate \channel. Standard \gls{ofdm} symbols, also called \textit{training} symbols, in the \wifi frame preamble enable this estimate through \cref{eq:freq-csi}. The resulting \gls{csi}, $\csi \in \complexm$, is the receiver's \emph{quantized} approximation of the channel response \channel for a given frame. After computing \csi, the receiver equalizes the remaining data symbols in the frame and recovers the transmitted symbols by computing
\begin{equation}
    \fsymbolrec = \csi^{-1} \odot \yfsymbol
    \label{eq:freq-csi}
\end{equation}
for successive values of $n$. When the receiver has $L>1$ antennas spaced at least half a wavelength apart, we generalize \cref{eq:freq-channel,eq:freq-csi} by deriving $\yfsymbolant$ for each receiving antenna $l$, thereby measuring the \gls{csi} $\csi^{(l)}$ of several channels simultaneously from a single \wifi frame transmission.

\subsection{Related Work on Wi-Fi Sensing}\label{sec:related}

\wifi sensing emerged as a research field in 2011, following the release of the first public \gls{csi} extraction tool~\cite{halperinToolReleaseGathering2011} and the pioneering work of~\cite{adibSeeWallsWiFi2013}. Researchers have since developed numerous methods that use \wifi signals for environmental monitoring. We focus here on \gls{har}, the application most directly related to our contribution, while other representative applications include person detection and identification~\cite{caiTeachingRFSense2020,soltanaghaeiRobustPracticalWiFi2020}, fall detection~\cite{yangRethinkingFallDetection2023}, and device-free localization~\cite{duDevicefreeIndoorLocalization2023}.

Early works~\cite{gaoCSIbasedDevicefreeWireless2017,zouDeepSenseDevicefreeHuman2018,chenWiFiCSIBased2019} established that human activities induce measurable variations in \gls{csi}, motivating the use of deep learning models to capture the complex relationship between wireless channel measurements and activity labels. Subsequent approaches explored a variety of architectures, including autoencoders~\cite{gaoCSIbasedDevicefreeWireless2017}, recurrent neural networks~\cite{zouDeepSenseDevicefreeHuman2018}, and long short-term memory models~\cite{chenWiFiCSIBased2019}. Comprehensive surveys of \gls{csi}-based \wifi sensing can be found in~\cite{miaoWifiSensingTechniques2025}.

Despite strong \gls{id} performance, many of these methods exhibit limited robustness when deployed across different environments, \gls{ood} humans, or various hardware configurations. To address this limitation, several studies proposed preprocessing techniques aimed at extracting environment-invariant features before classification. Representative examples include dimensionality reduction through principal component analysis~\cite{wangDevicefreeHumanActivity2017}, extraction of physically meaningful features such as micro-Doppler signatures~\cite{meneghelloSHARPEnvironmentPerson2023}, and signal processing pipelines for denoising and feature enhancement~\cite{jiangEnvironmentIndependentDevice2018}. Few-shot learning has also emerged as a promising strategy for cross-domain adaptation, with encouraging results on limited activity sets~\cite{bahadoriReWiSReliableWifi2022,yinFewSenseScalableCrossdomain2024,shiEnvironmentrobustDevicefreeHuman2022}. However, recent benchmarks show that generalization across realistic deployment conditions remains an open challenge~\cite{cominelliExposingCSISystematic2023}.

Recent work applies transformer-based architectures to \gls{csi}-based \gls{har}~\cite{liTwostreamConvolutionAugmented2021}, using self-attention to model long-range temporal dependencies across subcarriers. Although these models achieve strong predictive performance, they remain primarily discriminative because they map \gls{csi} measurements directly to activity labels without explicitly modeling the underlying data distribution. They consequently offer limited interpretability and no explicit mechanism for estimating uncertainty or separating environment-specific factors from activity-related information. Adversarial domain adaptation~\cite{miaoWifiSensingTechniques2025} provides another route to greater robustness, but it typically requires target-domain data during training.

In contrast, our approach adopts \glspl{vae} to learn compact and structured latent representations directly from \gls{csi} measurements. This enables both uncertainty-aware modeling and post-hoc interpretability without requiring environment-specific supervision.

\subsection{Variational Auto-Encoders}\label{sec:vae}

We use generative models to analyze the dependencies within the \gls{csi} dataset. Given a set of target classes $\mathcal{C}_K$, these models estimate either the class-conditional density $p(\bm{x} \,|\, \mathcal{C}_k)$ of a sample $\bm{x}$ or the joint distribution $p(\bm{x}, \mathcal{C}_K)$~\cite{bishopPatternRecognitionMachine2006}.

The \gls{vae} is a widely used generative model because it makes relatively mild modeling assumptions and supports efficient optimization through backpropagation~\cite{kingmaAutoencodingVariationalBayes2022}. A \gls{vae} assumes a dataset $\bX = {\{\bx^{(i)}\}}_{i=1}^N$ containing $N$ independent and identically distributed (i.i.d.) samples of a random variable $X$, generated through a process that involves an unobserved continuous random variable $Z$. Because both the generative parameters $\bT^*$ and the latent variables $\bzi$ underlying the observations $\bxi$ are unobserved, the model introduces a variational recognition model and learns its parameters $\bm{\phi}$. A probabilistic \emph{encoder} $P(Z \,|\, X, \bm{\phi})$ approximates the intractable true posterior $P(Z \,|\, X)$ by producing a distribution over the latent values from which a datapoint $\bx$ could have arisen, while a probabilistic \emph{decoder} $P(X \,|\, Z)$ produces a distribution over the observations associated with each latent value $Z$.

The marginal likelihood is the sum of the marginal likelihoods of the individual datapoints:
\begin{equation}
    \log p(\bx^{(1)}, \ldots, \bx^{(N)} \mid \bm{\theta}) = \sum_{i=1}^N \log p(\bxi \,|\, \bm{\theta}) \, ,
\end{equation}%
We can rewrite each term as
\begin{equation}
    \log \pT(\bxi) = \mbox{KL}\left(p(\bz \mid \bxi, \bm{\phi}) \,\middle\|\, p(\bz \mid \bxi, \bm{\theta})\right) + \LB{}{\bxi} \, ,
    \label{eq:marginallikelihood}
\end{equation}%
where $\mbox{KL}(p(\bz \,|\, \bxi, \bm{\phi}) ~||~ p(\bz \,|\, \bxi, \bm{\theta}))$ is the non-negative \gls{kl} divergence from the approximate posterior to the true posterior, and $\LB{}{\bxi}$ is the variational \emph{lower bound} on the marginal likelihood of datapoint $i$, \ie,
\begin{equation}
    \log \pT (\bxi) \geq \LB{}{\bxi}
    =                     \Exp{p(\bz \mid \bxi, \bm{\phi})}{- \log p(\bz \mid \bxi, \bm{\phi}) + \log p(\bxi,\bz \mid \bm{\theta})} \, .
    \label{eq:lowerbound}
\end{equation} %
We can therefore rewrite \cref{eq:marginallikelihood,eq:lowerbound} as
\begin{equation}
    \LB{}{\bxi} = - \mbox{KL}\left(p(\bz \mid \bxi, \bm{\phi}) \;\middle\|\; p(\bz \mid \bm{\theta})\right) + \Exp{p(\bz \mid \bxi, \bm{\phi})}{\log p(\bxi \mid  \bz, \bm{\theta})} \, ,
    \label{eq:lowerbound2}
\end{equation} %
where the chosen approximate posterior $p(\bz \,|\, \bx, \bm{\phi})$ is obtained through the \textit{reparameterization trick}~\cite{kingmaAutoencodingVariationalBayes2022} using a differentiable transformation $g(\beps,\bx \,|\, \bm{\phi})$ with $\beps \sim p(\beps)$. We can often integrate the \gls{kl} divergence analytically, whereas we estimate the expected reconstruction term $\Exp{p(\bz \,|\, \bxi, \bm{\phi})}{\log p(\bxi \,|\,  \bz, \bm{\theta})}$ by sampling $L$ values:
\begin{equation}
    \LB{}{\bxi}
    \simeq - \mbox{KL}\left(p(\bz \mid \bxi, \bm{\phi}) \;\middle\|\; p(\bz \mid \bm{\theta})\right) + \frac{1}{L} \sum_{l=1}^L \log p\left(\bxi \,|\, \bzil, \bm{\theta}\right) \, ,
    \label{eq:estimator2}
\end{equation}%
where $\bzil = g(\bxi,\bepsl \mid \bm{\phi})$ with $\beps \sim p(\beps)$. \Cref{eq:estimator2} is an estimator of the variational lower bound and is therefore maximized. Equivalently, we minimize its negative as a loss function: the \gls{kl} divergence of the approximate posterior acts as a regularization term, while the negative expected log likelihood represents the reconstruction error.

%% file: parts/3_methodology.tex
The \name architecture uses \glspl{vae} to learn compact latent representations of \gls{csi} measurements collected by passive \wifi radars. It exploits the multi-antenna capabilities of modern \wifi receivers by processing each antenna's \gls{csi} with a separate \gls{vae} and combining the resulting latent representations for downstream classification. In this section, we describe the experimental testbed, data preprocessing steps, architectural details, hyperparameter optimization strategy, and the hypotheses tested by the activity-recognition and deployment-shift experiments.

\subsection{Experimental Testbed}\label{sec:testbed}

We evaluate \name on the public \gls{csi} dataset introduced in~\cite{cominelliExposingCSISystematic2023}, which was collected in a realistic indoor environment with two ASUS RT-AX86U devices configured as transmitter and receiver while a human subject performed predefined activities. The dataset contains \gls{csi} measurements from IEEE~802.11ax transmissions over \SI{160}{\mega\hertz} channels with 2048 \gls{ofdm} subcarriers, sampled at 150 frames per second for 80 seconds per activity. We focus on scenario \texttt{S1}, which records participant A performing the twelve activities in \cref{tab:activities} in a laboratory, and use sub-scenario \texttt{S1a}, recorded by collector~a, to train and evaluate the model.

\begin{table}
    \centering
    \caption{The twelve activities used in this study, with the labels from the original dataset~\cite{cominelliExposingCSISystematic2023}.}\label{tab:activities}
    \footnotesize
    \begin{tabular}{llll}
        \toprule
        A.\,Walk       & B.\,Run   & C.\,Jump       & D.\,Sit     \\
        E.\,Empty room & F.\,Stand & G.\,Wave hands & H.\,Clap    \\
        I.\,Lie down   & J.\,Wipe  & K.\,Squat      & L.\,Stretch \\
        \bottomrule
    \end{tabular}
\end{table}

\begin{figure}
    \centering
    \includegraphics[width=.4\textwidth]{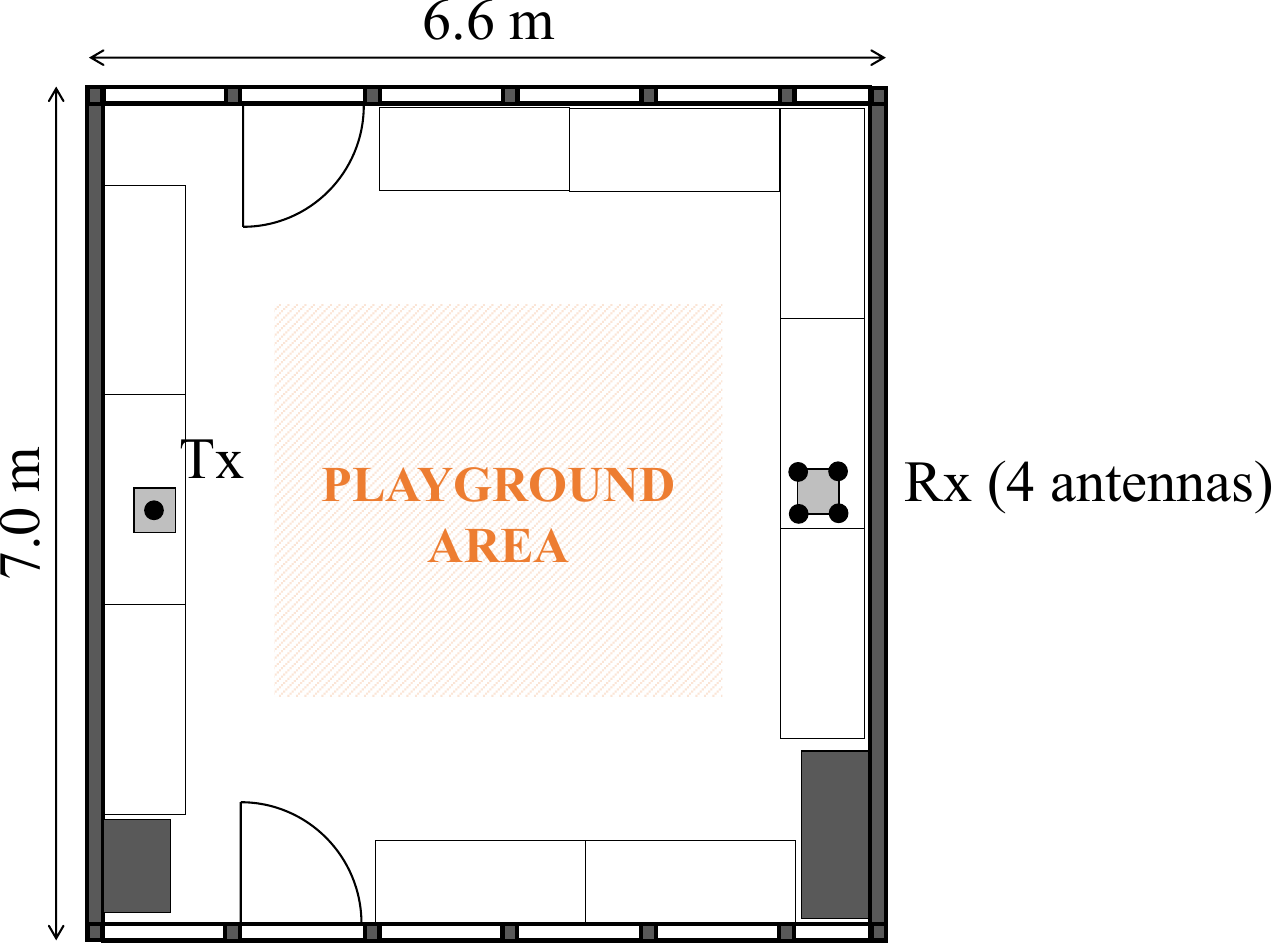}
    \caption{Layout of the experimental testbed. The highlighted area in the middle of the room indicates where the activities are performed.}\label{fig:layout}
\end{figure}

The activities take place in the central area of an approximately 46-square-metre room, with the transmitter on the west wall and the receiver on the east wall, as shown in \cref{fig:layout}. The receiver's four physically separated antennas compute the \gls{csi} of their corresponding channels simultaneously and independently. Participants perform static activities (\eg sitting and standing) at the center of the \emph{playground area} and dynamic activities (\eg walking and running) throughout the area.

\subsection{Data preprocessing}\label{sec:data-preprocessing}

In order to maintain the temporal structure of the \gls{csi} measurements while providing sufficient data for training, validation, and testing, we train each \gls{vae} on temporal windows obtained with a sliding-window procedure. Each window spans 3 seconds sampled at 150 frames per second, giving $T=450$ time steps, and consecutive windows begin 50 samples apart, which corresponds to a one-third-second stride. To reduce the data size while preserving the underlying physical channel response, we reduce the original \gls{csi} from 2048 to 256 subcarriers by discarding the upper half of the bandwidth and retaining every fourth remaining subcarrier. These reductions also make \name compatible with the older 802.11ac standard, which uses \SI{80}{\mega\hertz} channels. Each \gls{vae} therefore receives a \gls{csi}-magnitude window of shape $(T \times M)$, where $T=450$ is the number of time steps and $M=256$ is the number of subcarriers.

To prevent temporal data leakage caused by overlapping sliding windows while ensuring that evaluation sets remain representative of the full recording sessions, a deterministic block-wise temporal splitting procedure is applied to each activity recording. Because \gls{csi} is recorded continuously over extended time blocks, human activity execution naturally undergoes non-stationary variations, such as spatial drift or posture changes, causing early parts of a recording to differ physically from later parts. A simple single continuous split would create an unrepresentative test set containing only late-stage session dynamics, whereas random window shuffling would cause severe data leakage by placing overlapping adjacent windows into both training and testing sets. Instead, we partition each activity recording into four equal, contiguous temporal segments. Within each segment, the data is sliced sequentially without random shuffling into training, validation, and test sets using a 70:10:20 split ratio. The sliced subsets from all four segments are then aggregated for each activity class.

\subsection{Architecture}\label{sec:architecture}

\name has two modules. The first contains four \glspl{vae}, one per receiving antenna, which independently map the raw \gls{csi} magnitude to compressed multivariate-Gaussian latent representations. The second concatenates these four representations and uses a \gls{mlp} classifier to predict the target activity.

\begin{figure}
    \centering
    \includegraphics[width=.8\linewidth]{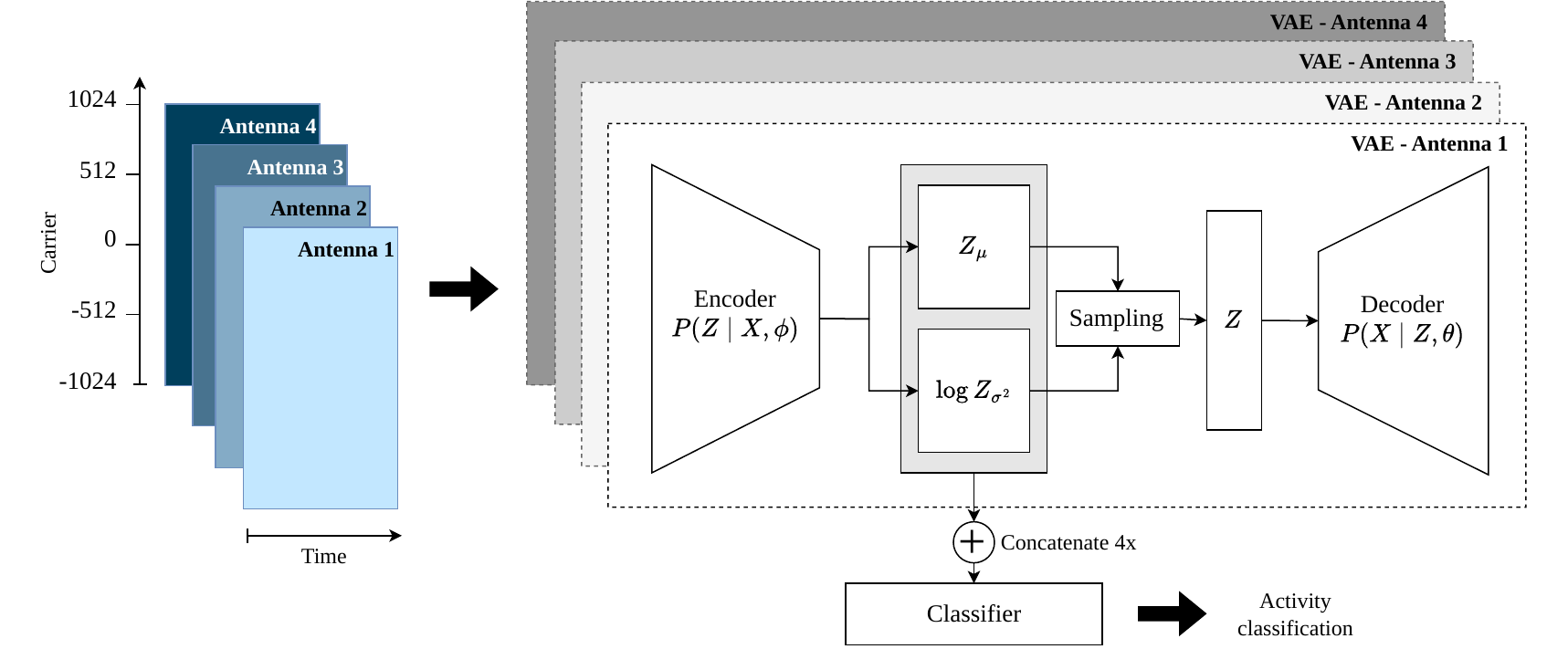}
    \caption{Scheme of the \name architecture. Each \gls{csi} magnitude spectrogram is processed by a separate \gls{vae} to extract a compressed latent representation. The latent representations from the four \glspl{vae} are then concatenated and fed to an \gls{mlp} classifier to predict the target activity.}\label{fig:architecture}
\end{figure}

Most \wifi sensing studies still use the single-antenna \gls{csi} extraction system introduced in~\cite{halperinToolReleaseGathering2011}. Current \wifi devices, however, usually have multiple antennas that can \textit{observe} several orthogonal wireless channels at the same time and thus provide a more informative description of the environment.

\begin{figure}
    \centering
    \includegraphics[width=\textwidth]{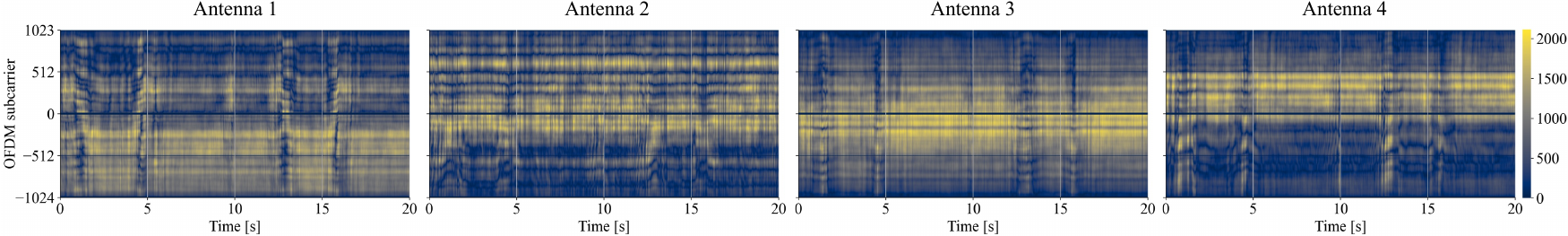}
    \caption{Time-frequency representation of the \acrshort{csi} magnitude collected by the four antennas of a single \wifi receiver while a person is walking inside the room. The four channels exhibit different properties because the antennas are physically spaced by more than half of the wavelength; thus, the indoor multipath effects on different antennas appear to be uncorrelated. We argue that combining the \acrshort{csi} collected simultaneously at different antennas can provide a more descriptive representation of the target activity.}\label{fig:walking-csi}
\end{figure}

Different antennas on the same receiver capture distinct channels directly in their \gls{csi} measurements, as illustrated in \cref{fig:walking-csi}. When the antennas are spaced by more than half a wavelength, complex indoor multipath effects make their \gls{csi} measurements uncorrelated.\ \name therefore processes the antennas jointly so that the classifier can use the complementary information they provide.

Although \gls{csi} is a complex vector that represents the wireless channel's frequency response (\cf \cref{sec:csi-details}), we use only its magnitude and leave the effect of phase information for future work. We interpret each sequence of \gls{csi}-magnitude measurements as a spectrogram and normalize its amplitude by the maximum value in the dataset.

For each of the \glspl{vae}, we assume that the prior over $Z$ is the centered isotropic multivariate Gaussian $p(\bz) = \mathcal{N}(\bz \mid \bzero, \bI)$. We use a factorized Gaussian decoder likelihood with fixed unit covariance,
\begin{equation}
    p(\bx \mid \bz, \bm{\theta}) = \mathcal{N}\left(\bx \,\middle|\, \hat{\bx}_{\bm{\theta}}(\bz), \bI\right) \, ,
    \label{eq:decoder-likelihood}
\end{equation}
where $\hat{\bx}_{\bm{\theta}}(\bz)$ is the decoder output and parametrizes the conditional mean; the decoder does not predict a covariance. Consequently, maximizing the log-likelihood term in \cref{eq:gaussian-estimator} is, up to an additive constant, equivalent to minimizing the squared reconstruction error. To approximate the intractable true posterior $p(Z \mid X)$, we assume it takes an approximate Gaussian form with diagonal covariance, so the variational approximate posterior is:
\begin{equation}
    \log p(\bz \mid \bxi, \bphi) = \log \mathcal{N}\left(\bz \,\middle|\, \bmu^{(i)}, \diag\left(\bsigma^{2 (i)}\right)\right) \, ,
\end{equation}%
where the mean $\bmu^{(i)}$ and standard deviation $\bsigma^{(i)}$ of the approximate posterior are the outputs of the encoder. Samples from the posterior are obtained via the reparameterization trick (\cf \cref{eq:estimator2}) as $\bzil = \bmu^{(i)} + \bsigma^{(i)} \odot \bepsl$, where $\bepsl \sim \mathcal{N}(\bzero,\bI)$. If the latent space comprises $J$ independent Gaussian dimensions, the resulting $\beta$-weighted \gls{elbo} estimator for datapoint $\bxi$ is:
\begin{equation}
    \LB{}{\bxi} \simeq \frac{\beta}{2} \sum_{j=1}^J \left[ 1 + \log{\left(\sigma_j^{(i)}\right)}^2 - {\left(\mu_j^{(i)}\right)}^2 - {\left(\sigma_j^{(i)}\right)}^2 \right] + \frac{1}{S} \sum_{s=1}^S \log p\left(\bxi \,\middle|\, \bz^{(i,s)}, \bm{\theta}\right) \, ,
    \label{eq:gaussian-estimator}
\end{equation}
where $S$ is the number of Monte Carlo samples drawn from the posterior and $\beta$ weights the \gls{kl} term. \Cref{eq:estimator2} presents the standard \gls{elbo}, which is recovered from \cref{eq:gaussian-estimator} when $\beta=1$; here, we use the $\beta$-weighted variant and optimize its maximum value $\beta_{\max}$ as described in \cref{sec:hyperparameters}.

Each antenna's \gls{vae} uses the symmetric convolutional encoder and decoder summarized in \cref{fig:architecture}. The encoder treats the $M$ subcarriers as channels and applies a stack of 2D convolutional layers only along the time axis, with kernel size and stride fixed at 1 along the subcarrier dimension, to compress the temporal dimension progressively. Batch normalization and a \gls{gelu} activation follow each convolutional layer, after which two independent linear heads produce the mean $\bmu$ and log-variance $\log\bsigma^2$ of the approximate posterior. The decoder mirrors this structure: a linear layer projects the sampled latent vector to the pre-flatten feature-map shape, and a stack of transposed convolutions then upsamples it along the time axis, using batch normalization and \gls{gelu} activations on every layer except the last. Optuna selects the convolutional layer count, kernel sizes, strides, and latent dimension $J$ (\cf \cref{sec:hyperparameters}).

\begin{table}
    \centering
    \caption{Structure of the per-antenna \gls{vae}. $M = 256$ is the number of subcarriers, $T = 450$ the window size, $C$ the number of convolutional layers (tuned via Optuna), $k_i$ and $s_i$ the kernel size and stride of the $i$-th layer along the time axis, and $J$ the latent dimension.}\label{tab:vae-params}
    \footnotesize
    \begin{tabular}{llll}
        \multicolumn{4}{l}{\textbf{Encoder}}                                                                   \\
        \toprule
        \textbf{Layer}             & \textbf{Kernel size/Nodes} & \textbf{Stride} & \textbf{Activation}        \\
        \midrule
        Input                      & $T \times M$               & \--             & \--                        \\
        Conv2D $\times C$          & $(k_i, 1) \times M$        & $(s_i, 1)$      & BatchNorm + GELU           \\
        Flatten                    & \--                        & \--             & \--                        \\
        Linear $\to \bmu$          & $J$                        & \--             & \--                        \\
        Linear $\to \log\bsigma^2$ & $J$                        & \--             & \--                        \\
        \midrule
        Latent space               & $2J$                       &                 &                            \\
        \bottomrule                                                                                            \\[1em]

        \multicolumn{4}{l}{\textbf{Decoder}}                                                                   \\
        \toprule
        \textbf{Layer}             & \textbf{Kernel size/Nodes} & \textbf{Stride} & \textbf{Activation}        \\
        \midrule
        Linear                     & flat dim                   & \--             & GELU                       \\
        Reshape                    & \--                        & \--             & \--                        \\
        ConvTranspose2D $\times C$ & $(k_i, 1) \times M$        & $(s_i, 1)$      & BatchNorm + GELU$^\dagger$ \\
        Output                     & $T \times M$               & \--             & \--                        \\
        \bottomrule
        \multicolumn{4}{l}{$^\dagger$ BatchNorm and GELU omitted on the final transposed convolution.}
    \end{tabular}
\end{table}

The fusion module concatenates the four \gls{vae} representations and passes them to the \gls{mlp} classifier, which predicts the target activity. We use a simple \gls{mlp} that operates on this low-dimensional representation to keep training fast and inference lightweight. Its input has size $8J$, where $J$ is the per-antenna latent dimension and the factor of 8 accounts for the mean and log-variance vectors from all four \glspl{vae}, while its 12 outputs correspond to the target activities. Optuna selects the number and width of the hidden layers, the activation function, the learning rate, and the batch size.

\subsection{Hyperparameters}\label{sec:hyperparameters}

The \name pipeline involves two sets of hyperparameters: those governing the per-antenna \glspl{vae}, and those governing the \gls{mlp} classifier. Given the number of interacting parameters and the cost of each training run, we use Optuna~\cite{akibaOptunaNextgenerationHyperparameter2019} with a \gls{tpe}~\cite{bergstraAlgorithmsHyperparameterOptimization2011} sampler to drive the search.

The \gls{vae} hyperparameters subject to optimization are: the learning rate $\eta \in [10^{-3}, 3 \times 10^{-2}]$ (log-uniform), the maximum \gls{kl} weight $\beta_{\max} \in [1.0, 20.0]$ (log-uniform), the batch size $B \in \{32, 64, 128\}$ (sampled via its base-2 exponent), and the convolutional layer specification, chosen categorically from the three candidates listed in \cref{tab:conv-specs}. Each specification defines a 3-layer encoder in which every entry gives the kernel size and stride along the time axis; the corresponding decoder uses the same specifications in reverse order. The \gls{mlp} classifier hyperparameters include the number of fusion layers $\in \{1, 2, 3\}$, fusion dropout rate $\in [0.0, 0.3]$ (step $0.1$), learning rate, and batch size. Rather than including $J$ in the Optuna search space, we run an independent study for each value of $J$, starting at 1 and increasing it across studies. This separation lets the \gls{tpe} sampler build a focused surrogate model for each latent dimension without conflating the effect of $J$ with the other hyperparameters.

Each study runs for at most $N_t=100$ trials and may stop early through Optuna's Expected Improvement to Median Ratio terminator~\cite{ishibashiStoppingCriterionBayesian2023}. We also stop the sequence of studies when the best median accuracy fails to improve by at least $\delta=0.02$ for two consecutive values of $J$, thereby avoiding computation beyond the point of diminishing returns. Because \gls{vae} training is sensitive to random initialization, we evaluate each Optuna trial with $N_s=10$ independent seeds. We randomly choose the seeds at the start, keep them fixed across every trial in a study, and use the median accuracy as the trial objective.

\begin{table}
    \centering
    \caption{Candidate convolutional layer specifications for the per-antenna \gls{vae} encoder. Each row defines a 3-layer stack; each entry is (kernel size, stride) along the time axis. The subcarrier dimension is never convolved (kernel size and stride fixed to 1).}\label{tab:conv-specs}
    \footnotesize
    \begin{tabular}{cccc}
        \toprule
        \textbf{Spec} & \textbf{Layer 1} & \textbf{Layer 2} & \textbf{Layer 3} \\
        \midrule
        A             & $(5, 5)$         & $(5, 5)$         & $(3, 3)$         \\
        B             & $(5, 5)$         & $(5, 5)$         & $(2, 2)$         \\
        C             & $(3, 3)$         & $(5, 5)$         & $(5, 5)$         \\
        \bottomrule
    \end{tabular}
\end{table}

\subsection{Experimental Hypotheses}\label{sec:hypothesis}

Our main hypothesis is that the per-antenna \glspl{vae} in \name provide a concise yet informative representation of the activities observed by the \wifi monitor. Each \gls{vae} receives raw \gls{csi} magnitudes without activity labels, so the reconstruction objective alone determines the structure of its latent space. We test this hypothesis quantitatively through the \gls{mlp} classifier's accuracy and qualitatively by inspecting the clusters formed by the latent representations.

A second hypothesis concerns the benefit of multi-antenna fusion. Since the four antennas of the monitor are spaced by more than half a wavelength, the received signals are generally uncorrelated due to indoor multipath effects (\cf \cref{fig:walking-csi}). Each per-antenna \gls{vae} therefore captures an uncorrelated view of the same activity, and concatenating the four latent representations gives the downstream classifier a richer, higher-dimensional description of the performed activity than any single antenna alone. We thus expect the \name architecture to achieve higher classification accuracy than a system relying on a single antenna's latent space.

A third hypothesis concerns deployment-shift detection. Because the variational representation describes each input through a distribution rather than a single deterministic feature vector, evidential classification heads should expose changes in the support and strength of the evidence when the participant, collector, recording session, or room differs from the S1a training domain. We therefore expect class-conditional uncertainty scores to rank shifted samples above held-out S1a samples, and we test whether this ranking remains useful at a calibrated 5\% in-distribution rejection level.

%% file: parts/4_results.tex
This section evaluates \name across latent-space sizes and architectures, examines how the best model represents the activities in its \gls{vae} latent spaces, and compares \name with independently optimized single-antenna models and their late fusion. To support reproducibility and further research on \wifi sensing, we publicly release the code\footnote{\url{https://github.com/lucactt/csi-vae}} and data.

\subsection{Architectures and Hyperparameters}\label{sec:results-architectures}

\begin{table}[t]
    \centering
    \caption{Hyperparameter search results across ten fixed random seeds. $\eta$ is the learning rate, $\beta_{\max}$ the maximum KL weight, $B$ the batch size, and \textit{Conv.} the convolutional specification from \cref{tab:conv-specs}.}\label{tab:results-architectures}
    \small
    \begin{tabular}{cccccccccc}
        \toprule
        \textbf{$J$} & \textbf{$\eta$} & \textbf{$\beta_{\max}$} & \textbf{$B$} & \textbf{Conv.} & \textbf{Fusion layers} & \textbf{Dropout} & \textbf{P25} & \textbf{Median} & \textbf{P75} \\
        \midrule
        1            & 0.0036          & 3.3460                  & 128          & C              & 2                      & 0.01             & 0.9031       & 0.9323          & 0.9411       \\
        2            & 0.0032          & 17.8439                 & 64           & C              & 2                      & 0.02             & 0.9594       & 0.9656          & 0.9703       \\
        3            & 0.0033          & 12.3290                 & 64           & A              & 1                      & 0.30             & 0.9526       & 0.9646          & 0.9688       \\
        4            & 0.0068          & 19.0895                 & 64           & C              & 1                      & 0.24             & 0.9547       & 0.9688          & 0.9750       \\
        5            & 0.0099          & 6.2654                  & 32           & C              & 1                      & 0.15             & 0.9521       & 0.9677          & 0.9688       \\
        6            & 0.0016          & 3.7391                  & 64           & B              & 1                      & 0.30             & 0.9563       & 0.9677          & 0.9724       \\
        \bottomrule
    \end{tabular}
\end{table}

The Optuna results in \cref{tab:results-architectures} show that $J=4$, which gives each antenna an eight-parameter latent representation, achieves the highest median accuracy. Its advantage over $J=2$ is marginal, however, which indicates that the additional latent parameters carry little information that helps the classifier distinguish the activities. Larger latent spaces also give the classifier more parameters to learn and process, increasing both the risk of overfitting and the computational cost.

We therefore define \name as the Optuna-optimized architecture with $J=2$, whose median accuracy is nearly identical to that of the larger latent spaces while requiring the classifier to process fewer parameters. The best-performing \name seed reaches 98.3\% test accuracy, compared with 70\% for the state-of-the-art SHARP method~\cite{meneghelloSHARPEnvironmentPerson2023} on the same dataset~\cite{cominelliExposingCSISystematic2023}.

\begin{figure}
    \centering
    \includegraphics[width=.5\textwidth]{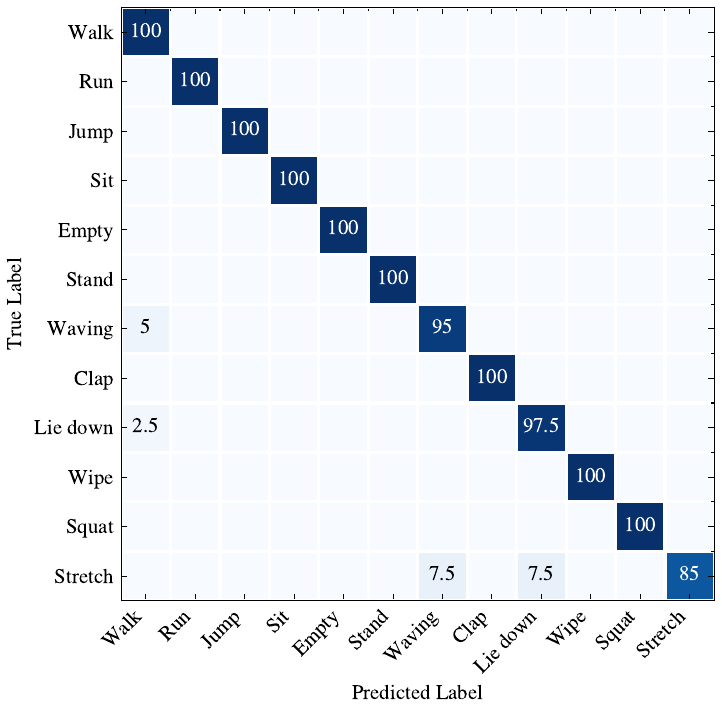}
    \caption{Confusion matrix for the best-performing \gls{vae} architecture with a latent space of 4 parameters.}\label{fig:best-confusion-matrix}
\end{figure}

The confusion matrix in \cref{fig:best-confusion-matrix} shows that the best model mainly confuses \textit{stretch}, \textit{waving}, and \textit{lie down}. Stretching and waving both involve arm movements, which may produce similar \gls{csi} patterns, whereas the confusion with lying down is less intuitive because that activity changes the body's overall position. We hypothesize that this larger positional change produces substantial channel variation that sometimes overlaps with the patterns generated by the other two activities.

\subsection{Visualization of the VAE Latent Spaces}\label{sec:results-vae}

\begin{figure}
    \centering
    \includegraphics[width=\textwidth]{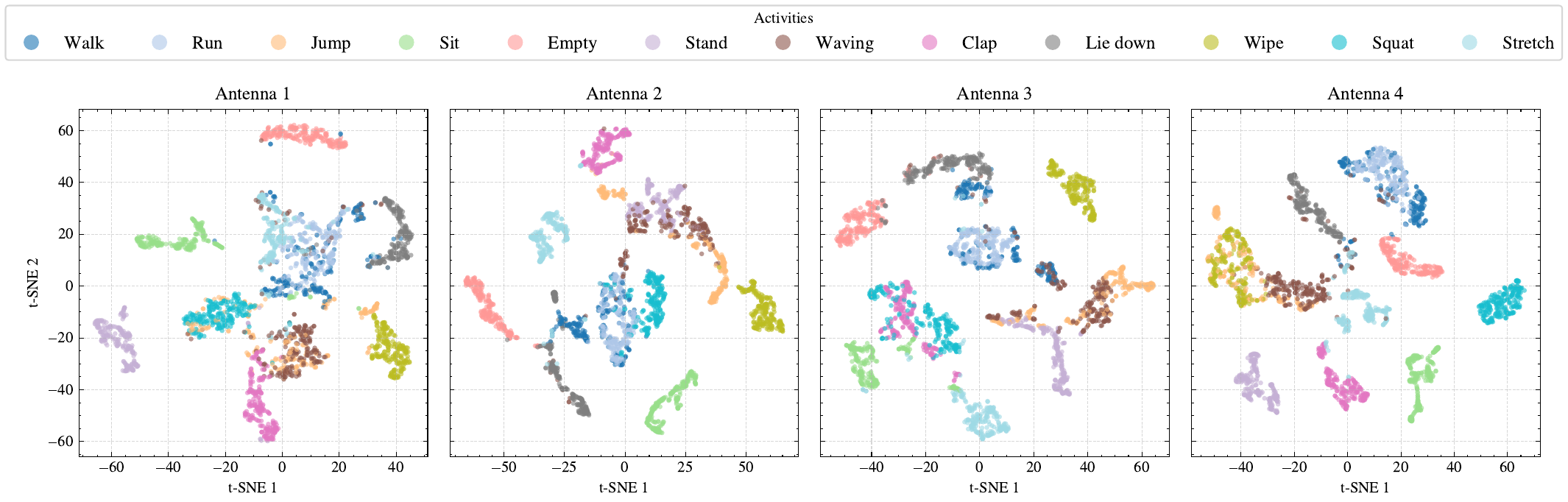}
    \caption{t-SNE projections of the latent spaces learned from different antennas, with $J=2$. Each original four-parameter point describes a bivariate Gaussian distribution, and the encoder maps each CSI window to one such point. Colors indicate the corresponding activities; clearer separation between activity clusters is associated with higher classification accuracy.}\label{fig:vae-latent-space}
\end{figure}

\Cref{fig:vae-latent-space} visualizes the latent spaces learned by the antenna-specific \glspl{vae} for the bivariate-Gaussian case described in \cref{sec:architecture}. We apply \gls{tsne}~\cite{vandermaatenVisualizingDataUsing2008} to each original latent space and project it onto two dimensions. Each colored point in the resulting $(\text{t-SNE}_1, \text{t-SNE}_2)$ plane represents a \gls{csi} window, with color indicating the activity performed during that window.

The antenna-specific \glspl{vae} separate several activity clusters despite receiving no activity labels, although the separation quality varies across antennas. Our preliminary tests found no clear disparity in the raw data collected by the four antennas, so data quality alone does not explain why some single-antenna latent spaces separate the activities more clearly than others for the four-parameter representation ($J=2$) in \cref{fig:vae-latent-space}. We instead expect single-antenna performance to depend strongly on the room geometry and experimental conditions because each antenna observes a different channel from its physical position. The projections also show meaningful overlaps: \emph{run} lies largely within \emph{walk}, while \emph{run} and \emph{jump}, as well as \emph{sit} and \emph{empty room}, overlap partially. The overlap between \emph{sit} and \emph{jump} may reflect the three-second window, within which brief segments of these activities can produce similar channel dynamics. These patterns confirm that different activities induce distinct but sometimes overlapping \gls{csi} distributions and support fusing multiple antennas when the classifier must distinguish many activities.

\subsection{Ablation Study}\label{sec:results-ablation}

\begin{figure}
    \centering
    \includegraphics[width=.5\textwidth]{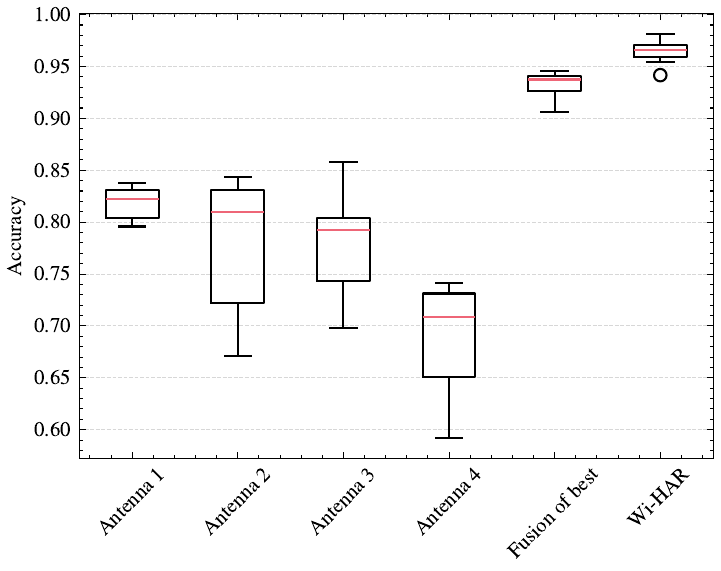}
    \caption{Ablation study comparing the best-performing \gls{vae} architecture with a latent space of 4 parameters against single-antenna architectures. ``Fusion of best'' trains a classifier on the latent representations of independently optimized single-antenna \glspl{vae}, whereas \name jointly optimizes the \glspl{vae} and classifier end-to-end.}\label{fig:ablation}
\end{figure}

\begin{table}
    \caption{Ablation study results. ``Fusion of best'' trains a classifier on the latent representations of independently optimized single-antenna \glspl{vae}, whereas \name jointly optimizes the \glspl{vae} and classifier end-to-end.}\label{tab:results-ablation}
    \center
    \begin{tabular}{@{}lccc@{}}
        \toprule
        \textbf{Name}  & \textbf{Median} & \textbf{P25}    & \textbf{P75}    \\
        \midrule
        Antenna 1      & 0.8146          & 0.8036          & 0.8313          \\
        Antenna 2      & 0.8063          & 0.7224          & 0.8313          \\
        Antenna 3      & 0.7875          & 0.7432          & 0.8036          \\
        Antenna 4      & 0.7083          & 0.6505          & 0.7313          \\
        Fusion of best & 0.9375          & 0.9266          & 0.9411          \\
        \textbf{\name} & \textbf{0.9646} & \textbf{0.9594} & \textbf{0.9703} \\
        \bottomrule
    \end{tabular}%
\end{table}

The ablation results show that every single-antenna architecture is less accurate and less stable than the fused alternatives. The best single-antenna model has a median accuracy below 85\%, and some runs fall to approximately 60\%, indicating strong sensitivity to room geometry and experimental conditions. The fused architectures maintain substantially higher accuracy across seeds, which supports our hypothesis that combining \gls{csi} measurements from multiple antennas improves \gls{har} performance.

\subsection{Interpreting the Learned Latent Space}\label{sec:interpretability}

We now examine what the compact variational representation encodes about activities, by training a shallow decision tree on the latent means and log-variances of the best-performing \name model. This analysis is deliberately separate from the main recognition results because it explains the learned features rather than defining another deployment output.

\begin{figure}
    \centering
    \begin{subfigure}{\textwidth}
        \centering
        \includegraphics[width=.8\textwidth]{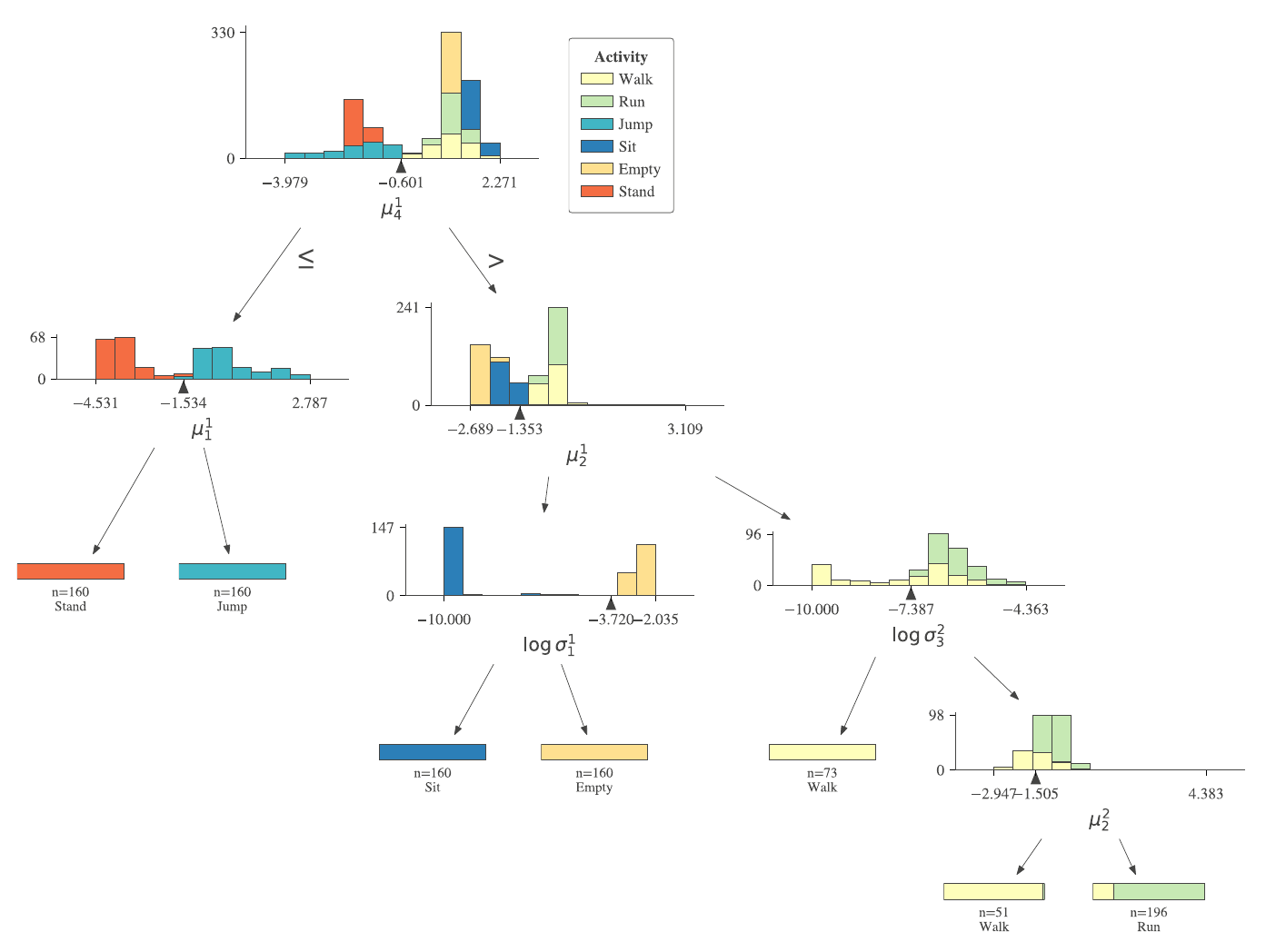}
        \caption{Representation of the decision tree. Each node shows the split condition, the number of samples in the node, and the predicted activity.}\label{fig:decision-tree-representation}
    \end{subfigure}
    \begin{subfigure}{\textwidth}
        \centering
        \includegraphics[width=.4\textwidth]{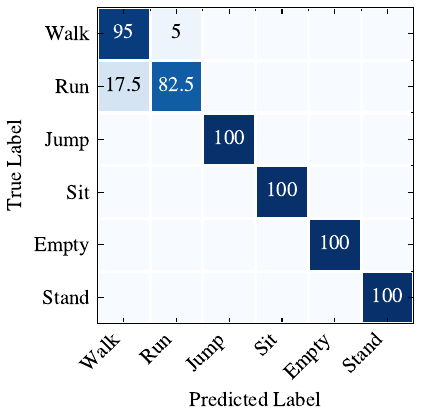}
        \caption{Confusion matrix for the decision tree.}\label{fig:decision-tree-confusion-matrix}
    \end{subfigure}
    \caption{Decision tree with depth of four for the first six activities and the $J=2$ representation.}\label{fig:decision-tree}
\end{figure}

We replace the standard classifier in \name with a shallow decision tree and restrict the analysis to the first six activities for readability. The tree operates directly on the means and log-variances of the approximate posterior distributions. We use $\mu^{i}_{j}$ and $\log \sigma^{i}_{j}$ to denote the $i$-th mean and log-variance from antenna $j$. Although the tree in \cref{fig:decision-tree} is limited to four splits, it reaches an overall test accuracy of 96.5\%.

The paths for \textit{walk} and \textit{run} are structurally similar, and the second latent mean of the second antenna, $\mu^2_2$, provides the main split between them. Their remaining confusion is consistent with the difficulty of separating related locomotion patterns from short \gls{csi} windows. The initial paths for \textit{stand} and \textit{jump} use $\mu^1_1$ as the main discriminant, which suggests that the first antenna captures a feature related to vertical position or posture. The log-variance of the same latent dimension, $\log \sigma^1_1$, helps distinguish \textit{sit} from \textit{empty room}; both create relatively static channels, but only the former includes a person in the monitored area.

Variables from several antennas appear in the final decision boundaries, which gives an interpretable counterpart to the ablation result in \cref{sec:results-ablation}: the fused classifier uses distinct views of the activity rather than relying on one dominant receiver channel. The tree also suggests different levels of compositional structure. Walking and running appear relatively atomic, whereas jumping may combine posture and translational components. This interpretation is consistent with~\cite{cottiCausallyInterpretableWiFi2026}, which studies atomic activity representations and their composition through neuro-symbolic event processing. These physical readings remain hypotheses, but the small latent space makes them explicit and testable.

%% file: parts/5_uncertainty.tex
High activity-recognition accuracy on S1a does not show whether the same classifier can recognize when its deployment conditions have changed. We therefore treat deployment-shift detection as a second output of the \name pipeline. The antenna-specific encoders and delayed-fusion backbone remain fixed, while alternative classification heads produce both activity predictions and scalar uncertainty scores.

\subsection{Uncertainty Heads}\label{sec:uncertainty-heads}

We compare the Softmax classifier used in \cref{sec:results} with \gls{edl}~\cite{sensoyEvidentialDeepLearning2018} and \gls{gen}~\cite{sensoyUncertaintyawareDeepClassifiers2020}. All three models use the same feature extractors and fusion backbone. We select the two evidential heads through the same Optuna workflow and validation split used for Softmax, with validation accuracy as the sole optimization objective; no shifted dataset contributes to model selection or calibration.

Both evidential heads produce non-negative evidence $e_k$ for each of the $K$ activity classes and define a Dirichlet distribution through
\begin{equation}
    \alpha_k=e_k+1, \qquad S=\sum_{j=1}^{K}\alpha_j,
    \qquad \widehat p_k=\frac{\alpha_k}{S}, \qquad u_{\mathrm{ev}}=\frac{K}{S}.
    \label{eq:evidential-summary}
\end{equation}
The predictive probability is the Dirichlet mean, while $u_{\mathrm{ev}}$ measures the absence of total evidence: it equals one when every evidence value is zero and decreases as the Dirichlet strength $S$ grows. The selected \gls{edl} head uses softplus evidence and the evidential cross-entropy Bayes risk with a 20-epoch regularization warm-up. The selected \gls{gen} head uses exponential evidence and an adversarial network with hidden width 128.

For the Softmax head, we use normalized predictive entropy,
\begin{equation}
    u_{\mathrm{SM}}=-\frac{1}{\log K}\sum_{k=1}^{K}p_k\log p_k,
    \label{eq:softmax-uncertainty-main}
\end{equation}
which also lies in $[0,1]$ but has a different interpretation from \cref{eq:evidential-summary}: it measures dispersion in one categorical distribution rather than inverse Dirichlet strength.

\subsection{Class-conditional Deployment-shift Score}\label{sec:shift-score}

The uncertainty distribution depends on the predicted activity, so a pooled reference can hide a shift that is clear within one predicted class. For model $m$ and predicted class $k$, we construct the training reference
\begin{equation}
    \mathcal{U}_{m,k}=\{u_i:\widehat y_i=k,\ i\in\texttt{TRAIN}\},
    \qquad n_{m,k}=|\mathcal{U}_{m,k}|,
\end{equation}
and evaluate a test uncertainty $u$ with the mid-rank empirical \gls{cdf} $\widehat F_{m,k}(u)$. We then define the two-sided anomaly score
\begin{equation}
    A_m(u,k)=-\log\!\left[
        \max\!\left(
        2\min\{\widehat F_{m,k}(u),1-\widehat F_{m,k}(u)\},
        \frac{1}{n_{m,k}+1}
        \right)\right].
    \label{eq:shift-score-main}
\end{equation}
A large value means that the uncertainty is unusual for S1a samples assigned to the same predicted class. Both tails matter because a deployment change can produce unusually high uncertainty or atypically confident predictions. The score measures disagreement with the class-conditional training reference; it is not a posterior probability that a sample is out of distribution.

\subsection{Evaluation Protocol}\label{sec:shift-protocol}

The held-out S1a partition supplies the in-distribution samples for every comparison, while each shifted dataset is evaluated separately. The benchmark contains four controlled classes of deployment change: S1b\--S1c change only the collector position; S2a\--S2c replace participant A with participant B;\@ S4a\--S4c repeat participant A in the same laboratory a few hours later; and S6a\--S6c move participant A to an office on the second collection day. Each comparison contains 480 held-out S1a samples and 480 samples from the corresponding shifted dataset. No shifted scenario is used for training, hyperparameter selection, empirical-\gls{cdf} construction, or threshold calibration.

We first report \gls{roc}\--\gls{auc}, which measures the probability that a randomly selected shifted sample receives a larger anomaly score than a randomly selected S1a sample. We then evaluate an explicit operating point with split-conformal calibration. Within every predicted class, 80\% of the training samples form the uncertainty reference and 20\% form a calibration set. For a sample $x$, we use its uncertainty and predicted class to compute the anomaly score from \cref{eq:shift-score-main}, $A(x)=A_m(u_m(x),\widehat y_m(x))$. Thus, high uncertainty relative to the predicted-class reference yields a high score, as can atypically low uncertainty because the score is two-sided. For calibration anomaly scores $A_1,\ldots,A_{n_{\mathrm{cal}}}$ and test anomaly score $A(x)$, the in-distribution compatibility value is
\begin{equation}
    p_{\mathrm{ID}}(x)=
    \frac{1+\sum_{j=1}^{n_{\mathrm{cal}}}\mathbb{I}[A_j\geq A(x)]}
    {n_{\mathrm{cal}}+1}.
    \label{eq:conformal-id-main}
\end{equation}
We reject a sample when $p_{\mathrm{ID}}(x)\leq0.05$ and repeat the class-stratified split 50 times, leaving held-out S1a untouched for measurement of the realized false-positive rate.

\subsection{Shift-detection Results}\label{sec:shift-results}

\begin{table}
    \centering
    \caption{\gls{roc}\--\gls{auc} of the predicted-class-conditional two-sided uncertainty score. S1a is the in-distribution reference; each remaining scenario is treated independently as the positive (shifted) class. Bold values indicate the highest AUC per scenario. Blue-highlighted scenarios are the most similar to S1a; orange-highlighted scenarios are the most dissimilar.}\label{tab:ood-auc}
    \small
    \begin{tabular}{lccc}
        \toprule
        \textbf{Scenario} & \textbf{Softmax} & \textbf{\gls{edl}} & \textbf{\gls{gen}} \\
        \midrule
        S1b               & 0.918            & \textbf{0.970}     & 0.968              \\
        S1c               & 0.327            & \textbf{0.978}     & 0.901              \\
        \rowcolor{SkyBlue} S2a               & 0.512            & \textbf{0.693}     & 0.571              \\
        S2b               & 0.648            & \textbf{0.970}     & 0.965              \\
        S2c               & 0.541            & \textbf{0.974}     & 0.956              \\
        \rowcolor{SkyBlue} S4a               & 0.702            & 0.790              & \textbf{0.814}     \\
        S4b               & 0.927            & \textbf{0.971}     & 0.883              \\
        S4c               & 0.736            & \textbf{0.972}     & 0.756              \\
        \rowcolor{Dandelion} S6a               & 0.664            & \textbf{0.905}     & 0.887              \\
        \rowcolor{Dandelion} S6b               & 0.606            & 0.672              & \textbf{0.796}     \\
        \rowcolor{Dandelion} S6c               & 0.696            & \textbf{0.972}     & 0.535              \\
        \midrule
        Macro average     & 0.662            & \textbf{0.897}     & 0.821              \\
        \bottomrule
    \end{tabular}
\end{table}

\Cref{tab:ood-auc} shows that \gls{edl} ranks shifted samples more consistently than the other models. It achieves the highest AUC in nine of eleven scenarios and a macro average of 0.897, compared with 0.821 for \gls{gen} and 0.662 for Softmax. The result is obtained even though validation accuracy, rather than shifted-data performance, selected the evidential heads.

Collector position alone causes a strong detectable shift. With the same participant, room, day, activities, and recording session, \gls{edl} reaches AUCs of 0.970 and 0.978 on S1b and S1c. Participant changes are collector-dependent: S2a is difficult for all three heads, while \gls{edl} reaches 0.970 and 0.974 on S2b and S2c. The later same-day session also changes the uncertainty distributions, with \gls{edl} AUCs from 0.790 to 0.972. In the office, performance again depends on the collector: \gls{edl} is strongest on S6a and S6c, whereas \gls{gen} is strongest on S6b. These differences show that the same physical change does not produce a uniform scalar response across receiver positions.

\begin{table}
    \centering
    \caption{Performance at the nominal 5\% split-conformal rejection level. Values are averaged over 50 class-stratified splits; shift-detection rates are macro-averages across the eleven shifted datasets, and intervals contain the central 90\% of results.}\label{tab:ood-conformal-operating-point}
    \small
    \begin{tabular}{lcccc}
        \toprule
        \textbf{Model} & \textbf{S1a false positives} & \textbf{90\% interval} & \textbf{Mean shift detection} & \textbf{90\% interval} \\
        \midrule
        Softmax        & 5.57\%                       & [4.58\%, 6.99\%]       & 22.54\%                       & [19.52\%, 30.26\%]     \\
        \gls{edl}      & 5.21\%                       & [4.03\%, 6.04\%]       & \textbf{82.51\%}              & [81.95\%, 82.94\%]     \\
        \gls{gen}      & 3.91\%                       & [3.13\%, 5.53\%]       & 65.12\%                       & [62.84\%, 66.45\%]     \\
        \bottomrule
    \end{tabular}
\end{table}

The calibrated operating point in \cref{tab:ood-conformal-operating-point} agrees with the AUC ranking. \Gls{edl} realizes a 5.21\% S1a false-positive rate, close to the nominal 5\% target, and achieves a macro-average shift-detection rate of 82.51\% across the eleven shifted datasets. \Gls{gen} achieves 65.12\%, while Softmax achieves 22.54\%. The evidential advantage therefore persists when the detector is required to operate at a controlled in-distribution rejection level.

\subsection{Comparison of the Uncertainty Heads}\label{sec:edl-gen-comparison}

Several scenarios give similar AUCs for \gls{edl} and \gls{gen}, so we use a paired Bayesian bootstrap with a \gls{rope} of $[-0.02,0.02]$ AUC.\@ The same Dirichlet bootstrap weights are applied to both models in every draw, preserving the paired S1a and shifted observations. Differences smaller than two AUC points are treated as negligible; this margin is of the same order as the scenario-level sampling uncertainty and corresponds to two additional correctly ordered in-distribution-shifted pairs per 100 random pairs.

\begin{figure}
    \centering
    \begin{subfigure}{0.444\textwidth}
        \includegraphics[width=\textwidth]{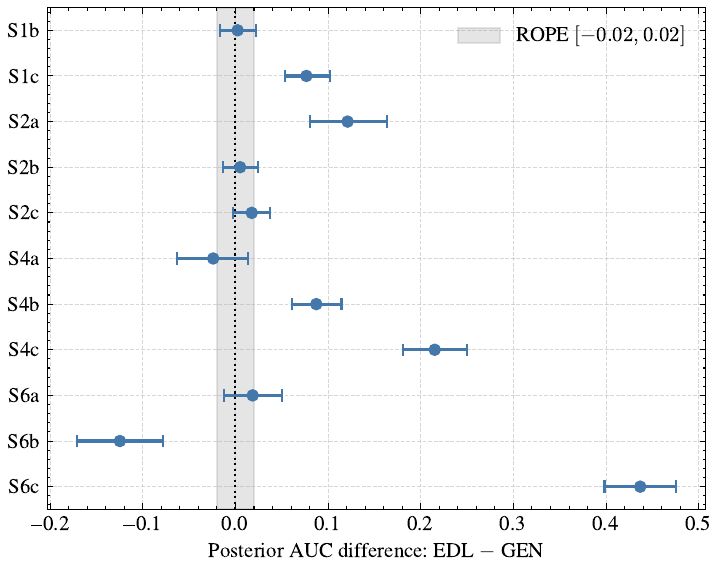}
        \caption{Posterior median and 95\% credible interval of $\mathrm{AUC}_{\mathrm{EDL}}-\mathrm{AUC}_{\mathrm{GEN}}$.}\label{fig:edl-gen-bayesian-rope-posterior}
    \end{subfigure}
    \begin{subfigure}{0.536\textwidth}
        \includegraphics[width=\textwidth]{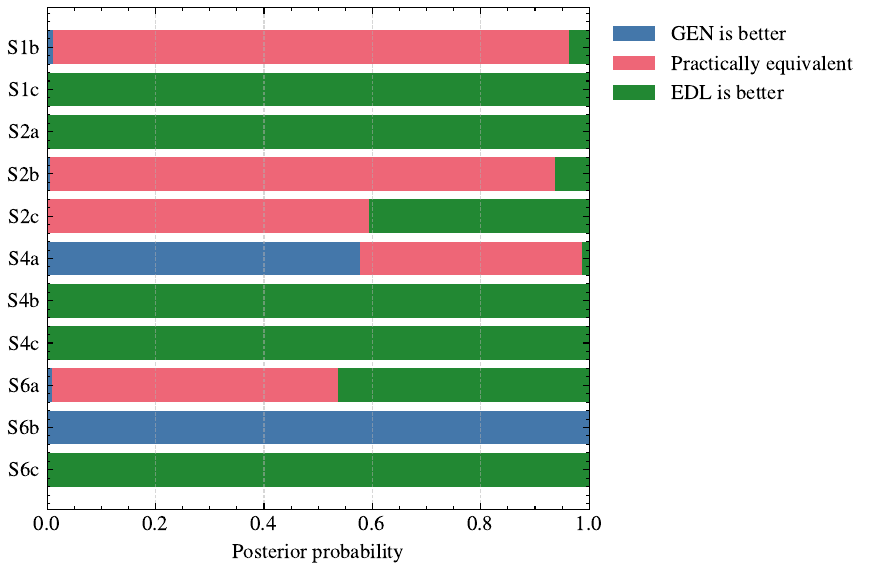}
        \caption{Probabilities of a meaningful \gls{gen} or \gls{edl} advantage, or practical equivalence.}\label{fig:edl-gen-bayesian-rope-probabilities}
    \end{subfigure}
    \caption{Paired Bayesian-bootstrap comparison of \gls{edl} and \gls{gen}.\ \cref{fig:edl-gen-bayesian-rope-posterior} shows the posterior median and 95\% credible interval of $\mathrm{AUC}_{\mathrm{EDL}}-\mathrm{AUC}_{\mathrm{GEN}}$; \cref{fig:edl-gen-bayesian-rope-probabilities} gives the probabilities of a meaningful \gls{gen} advantage, practical equivalence, and a meaningful \gls{edl} advantage under the $\pm0.02$ \gls{rope}.}\label{fig:edl-gen-bayesian-rope}
\end{figure}

\begin{table}
    \centering
    \caption{Aggregate Bayesian comparison for $\Delta=\mathrm{AUC}_{\mathrm{EDL}}-\mathrm{AUC}_{\mathrm{GEN}}$ under the $\pm0.02$ \gls{rope}.}\label{tab:edl-gen-rope-aggregate}
    \small
    \begin{tabular}{lccccc}
        \toprule
        \textbf{Estimand}      & \textbf{Mean} & \textbf{95\% credible interval} & \textbf{$P(\mathrm{GEN})$} & \textbf{$P(\mathrm{ROPE})$} & \textbf{$P(\mathrm{EDL})$} \\
        \midrule
        Fixed eleven scenarios & 0.0757        & [0.0581, 0.0934]                & 0.000                      & 0.000                       & 1.000                      \\
        Scenario bootstrap     & 0.0764        & [0.0069, 0.1721]                & 0.003                      & 0.059                       & 0.938                      \\
        \bottomrule
    \end{tabular}
\end{table}

The scenario-level posteriors in \cref{fig:edl-gen-bayesian-rope} show practical equivalence for S1b and S2b and close results for S2c and S6a. They favor \gls{edl} decisively on S1c, S2a, S4b, S4c, and S6c, while S6b decisively favors \gls{gen}. The asymmetric failures matter: the largest \gls{edl} gains are 0.216 AUC on S4c and 0.437 on S6c, whereas the largest \gls{gen} gain is 0.124 on S6b.

At the benchmark level, \cref{tab:edl-gen-rope-aggregate} gives a mean \gls{edl} advantage of 0.0757 AUC with a 95\% credible interval of [0.0581, 0.0934]. A second bootstrap over the mixture of deployment scenarios gives a 0.938 posterior probability that \gls{edl} retains an advantage larger than 0.02 AUC.\@ The two evidential heads are therefore interchangeable under several specific shifts, but the benchmark as a whole supports \gls{edl} as the more reliable detector.

%% file: parts/6_conclusions.tex
This paper presents \name as a compact multi-antenna framework for both activity recognition and deployment-shift detection from passive \wifi radar measurements. Four antenna-specific \glspl{vae} learn distributional representations from raw \gls{csi} magnitudes, and delayed fusion combines their means and log-variances. On twelve activities from the public benchmark, the selected model reaches a median accuracy of 96.6\%, while independently optimized single-antenna models remain below 82\%. The ablation study therefore identifies multi-antenna fusion as a central source of recognition performance.

The uncertainty study extends the system beyond closed-set recognition. Using the same latent backbone, \gls{edl} reaches a macro-average \gls{roc}\--\gls{auc} of 0.897 across eleven datasets that change collector position, participant, recording session, or room, and it detects 82.51\% of shifted samples at a realized 5.21\% S1a rejection rate. A paired Bayesian comparison shows that \gls{edl} and \gls{gen} are practically equivalent under several individual shifts, while the full benchmark favors \gls{edl} by about 0.076 AUC.\@ Detection performance nevertheless varies substantially across scenarios, showing that the same type of physical change does not produce a uniform uncertainty response across receiver positions. These results show that deployment changes can be detected without using shifted samples during model selection or calibration, although the scalar score cannot identify which physical factor caused the change.

The separate interpretability analysis shows why the compact representation is useful beyond predictive accuracy. A depth-four decision tree reaches 96.5\% accuracy on six activities and uses means and log-variances from several antennas, with individual splits suggesting relationships with locomotion, posture, and human presence.
Future work should test these interpretations through controlled physical interventions, incorporate phase information, and evaluate recognition after a shift has been detected. The broader challenge is to combine shift awareness with adaptation so that a \wifi sensing system can both flag an unfamiliar deployment and recover reliable activity predictions across people, receiver placements, and environments.

%% file: references.bib
@inproceedings{cominelliExposingCSISystematic2023,
  author     = {Cominelli, Marco and Gringoli, Francesco and Restuccia, Francesco},
  booktitle  = {2023 {{IEEE International Conference}} on {{Pervasive Computing}} and {{Communications}} ({{PerCom}})},
  doi        = {10.1109/PERCOM56429.2023.10099368},
  issn       = {2474-249X},
  langid     = {english},
  month      = mar,
  pages      = {81--90},
  shorttitle = {Exposing the {{CSI}}},
  title      = {Exposing the {{CSI}}: A Systematic Investigation of {{CSI-based}} Wi-Fi Sensing Capabilities and Limitations},
  urldate    = {2026-01-25},
  year       = 2023
}

@misc{kingmaAutoencodingVariationalBayes2022,
  archiveprefix = {arXiv},
  author        = {Kingma, Diederik P. and Welling, Max},
  doi           = {10.48550/arXiv.1312.6114},
  eprint        = {1312.6114},
  langid        = {english},
  month         = dec,
  number        = {arXiv:1312.6114},
  primaryclass  = {stat},
  publisher     = {arXiv},
  title         = {Auto-Encoding Variational Bayes},
  urldate       = {2026-02-19},
  year          = 2022
}

@inproceedings{cominelliAccuratePassiveRadar2023,
  author    = {Cominelli, Marco and Gringoli, Francesco and Kaplan, Lance M. and Srivastava, Mani B. and Cerutti, Federico},
  booktitle = {2023 26th {{International Conference}} on {{Information Fusion}} ({{FUSION}})},
  doi       = {10.23919/FUSION52260.2023.10224098},
  langid    = {english},
  month     = jun,
  pages     = {1--8},
  title     = {Accurate Passive Radar via an Uncertainty-Aware Fusion of Wi-Fi Sensing Data},
  urldate   = {2026-01-08},
  year      = 2023
}

@article{liPassiveWiFiRadar2021,
  author  = {Li, Wenda and Piechocki, Robert J. and Woodbridge, Karl and Tang, Chong and Chetty, Kevin},
  doi     = {10.1109/TGRS.2020.3006387},
  issn    = {1558-0644},
  journal = {IEEE Transactions on Geoscience and Remote Sensing},
  langid  = {english},
  month   = mar,
  number  = {3},
  pages   = {1986--1998},
  title   = {Passive {{WiFi}} Radar for Human Sensing Using a Stand-Alone Access Point},
  urldate = {2026-03-28},
  volume  = {59},
  year    = 2021
}

@article{storrerIndoorTrackingMultiple2021,
  author  = {Storrer, Laurent and Yildirim, Hasan Can and Crauwels, Morgane and Copa, Evert I. Pocoma and Pollin, Sofie and Louveaux, J{\'e}r{\^o}me and De Doncker, Philippe and Horlin, Fran{\c c}ois},
  doi     = {10.1109/JSEN.2021.3095675},
  issn    = {1558-1748},
  journal = {IEEE Sensors Journal},
  langid  = {english},
  month   = sep,
  number  = {18},
  pages   = {20462--20474},
  title   = {Indoor Tracking of Multiple Individuals with an 802.11ax Wi-Fi-Based Multi-Antenna Passive Radar},
  urldate = {2026-03-28},
  volume  = {21},
  year    = 2021
}

@article{seyfiogluDeepConvolutionalAutoencoder2018,
  author  = {Seyfio{\u g}lu, Mehmet Sayg{\i}n and {\"O}zbayo{\u g}lu, Ahmet Murat and G{\"u}rb{\"u}z, Sevgi Zubeyde},
  doi     = {10.1109/TAES.2018.2799758},
  issn    = {1557-9603},
  journal = {IEEE Transactions on Aerospace and Electronic Systems},
  langid  = {english},
  month   = aug,
  number  = {4},
  pages   = {1709--1723},
  title   = {Deep Convolutional Autoencoder for Radar-Based Classification of Similar Aided and Unaided Human Activities},
  urldate = {2026-03-28},
  volume  = {54},
  year    = 2018
}

@book{bishopPatternRecognitionMachine2006,
  author    = {Bishop, Christopher M and Nasrabadi, Nasser M},
  langid    = {english},
  publisher = {Springer},
  title     = {Pattern Recognition and Machine Learning},
  volume    = {4},
  year      = 2006
}

@article{halperinToolReleaseGathering2011,
  author     = {Halperin, Daniel and Hu, Wenjun and Sheth, Anmol and Wetherall, David},
  doi        = {10.1145/1925861.1925870},
  issn       = {0146-4833},
  journal    = {SIGCOMM Comput. Commun. Rev.},
  langid     = {english},
  month      = jan,
  number     = {1},
  pages      = {53},
  shorttitle = {Tool Release},
  title      = {Tool Release: Gathering 802.11n Traces with Channel State Information},
  urldate    = {2026-03-28},
  volume     = {41},
  year       = 2011
}

@article{sensoyUncertaintyawareDeepClassifiers2020,
  author    = {Sensoy, Murat and Kaplan, Lance and Cerutti, Federico and Saleki, Maryam},
  copyright = {Copyright (c) 2020 Association for the Advancement of Artificial Intelligence},
  doi       = {10.1609/aaai.v34i04.6015},
  issn      = {2374-3468},
  journal   = {Proceedings of the AAAI Conference on Artificial Intelligence},
  langid    = {english},
  month     = apr,
  number    = {4},
  pages     = {5621--5627},
  title     = {Uncertainty-Aware Deep Classifiers Using Generative Models},
  urldate   = {2026-03-28},
  volume    = {34},
  year      = 2020
}

@inproceedings{adibSeeWallsWiFi2013,
  address   = {Hong Kong China},
  author    = {Adib, Fadel and Katabi, Dina},
  booktitle = {{{SIGCOMM}}'13: {{ACM SIGCOMM}} 2013 {{Conference}}},
  doi       = {10.1145/2486001.2486039},
  isbn      = {978-1-4503-2056-6},
  langid    = {english},
  month     = aug,
  pages     = {75--86},
  publisher = {ACM},
  title     = {See through Walls with {{WiFi}}!},
  urldate   = {2026-03-28},
  year      = 2013
}

@article{caiTeachingRFSense2020,
  author  = {Cai, Hong and Korany, Belal and Karanam, Chitra R. and Mostofi, Yasamin},
  doi     = {10.1145/3432224},
  journal = {Proc. ACM Interact. Mob. Wearable Ubiquitous Technol.},
  langid  = {english},
  month   = dec,
  number  = {4},
  pages   = {120:1--120:22},
  title   = {Teaching {{RF}} to Sense without {{RF}} Training Measurements},
  urldate = {2026-03-28},
  volume  = {4},
  year    = 2020
}

@inproceedings{soltanaghaeiRobustPracticalWiFi2020,
  address   = {New York, NY, USA},
  author    = {Soltanaghaei, Elahe and Sharma, Rahul Anand and Wang, Zehao and Chittilappilly, Adarsh and Luong, Anh and Giler, Eric and Hall, Katie and Elias, Steve and Rowe, Anthony},
  booktitle = {Proceedings of the 7th {{ACM International Conference}} on {{Systems}} for {{Energy-Efficient Buildings}}, {{Cities}}, and {{Transportation}}},
  doi       = {10.1145/3408308.3427983},
  isbn      = {978-1-4503-8061-4},
  langid    = {english},
  month     = nov,
  pages     = {150--159},
  publisher = {Association for Computing Machinery},
  series    = {{{BuildSys}} '20},
  title     = {Robust and Practical {{WiFi}} Human Sensing Using On-Device Learning with a Domain Adaptive Model},
  urldate   = {2026-03-28},
  year      = 2020
}

@article{yangRethinkingFallDetection2023,
  author  = {Yang, Zheng and Zhang, Yi and Zhang, Qian},
  doi     = {10.1109/TMC.2022.3188779},
  issn    = {1558-0660},
  journal = {IEEE Transactions on Mobile Computing},
  langid  = {english},
  month   = oct,
  number  = {10},
  pages   = {6126--6143},
  title   = {Rethinking Fall Detection with Wi-Fi},
  urldate = {2026-03-28},
  volume  = {22},
  year    = 2023
}

@article{duDevicefreeIndoorLocalization2023,
  author  = {Du, Liufeng and Tian, Xiyan and Zhang, Linghua and Shang, Shaoru and Ning, Xin},
  doi     = {10.1109/ACCESS.2023.3263277},
  issn    = {2169-3536},
  journal = {IEEE Access},
  langid  = {english},
  pages   = {32548--32563},
  title   = {Device-Free Indoor Localization Based on Multidimensional {{CSI}} Features Classification},
  urldate = {2026-03-28},
  volume  = {11},
  year    = 2023
}

@article{gaoCSIbasedDevicefreeWireless2017,
  author  = {Gao, Qinhua and Wang, Jie and Ma, Xiaorui and Feng, Xueyan and Wang, Hongyu},
  doi     = {10.1109/TVT.2017.2737553},
  issn    = {1939-9359},
  journal = {IEEE Transactions on Vehicular Technology},
  langid  = {english},
  month   = nov,
  number  = {11},
  pages   = {10346--10356},
  title   = {{{CSI-based}} Device-Free Wireless Localization and Activity Recognition Using Radio Image Features},
  urldate = {2026-03-28},
  volume  = {66},
  year    = 2017
}

@inproceedings{zouDeepSenseDevicefreeHuman2018,
  author     = {Zou, Han and Zhou, Yuxun and Yang, Jianfei and Jiang, Hao and Xie, Lihua and Spanos, Costas J.},
  booktitle  = {2018 {{IEEE International Conference}} on {{Communications}} ({{ICC}})},
  doi        = {10.1109/ICC.2018.8422895},
  issn       = {1938-1883},
  langid     = {english},
  month      = may,
  pages      = {1--6},
  shorttitle = {{{DeepSense}}},
  title      = {{{DeepSense}}: Device-Free Human Activity Recognition via Autoencoder Long-Term Recurrent Convolutional Network},
  urldate    = {2026-03-28},
  year       = 2018
}

@article{chenWiFiCSIBased2019,
  author  = {Chen, Zhenghua and Zhang, Le and Jiang, Chaoyang and Cao, Zhiguang and Cui, Wei},
  doi     = {10.1109/TMC.2018.2878233},
  issn    = {1558-0660},
  journal = {IEEE Transactions on Mobile Computing},
  langid  = {english},
  month   = nov,
  number  = {11},
  pages   = {2714--2724},
  title   = {{{WiFi CSI}} Based Passive Human Activity Recognition Using Attention Based {{BLSTM}}},
  urldate = {2026-03-28},
  volume  = {18},
  year    = 2019
}

@article{wangDevicefreeHumanActivity2017,
  author  = {Wang, Wei and Liu, Alex X. and Shahzad, Muhammad and Ling, Kang and Lu, Sanglu},
  doi     = {10.1109/JSAC.2017.2679658},
  issn    = {1558-0008},
  journal = {IEEE Journal on Selected Areas in Communications},
  langid  = {english},
  month   = may,
  number  = {5},
  pages   = {1118--1131},
  title   = {Device-Free Human Activity Recognition Using Commercial {{WiFi}} Devices},
  urldate = {2026-03-28},
  volume  = {35},
  year    = 2017
}

@inproceedings{jiangEnvironmentIndependentDevice2018,
  address   = {New York, NY, USA},
  author    = {Jiang, Wenjun and Miao, Chenglin and Ma, Fenglong and Yao, Shuochao and Wang, Yaqing and Yuan, Ye and Xue, Hongfei and Song, Chen and Ma, Xin and Koutsonikolas, Dimitrios and Xu, Wenyao and Su, Lu},
  booktitle = {Proceedings of the 24th {{Annual International Conference}} on {{Mobile Computing}} and {{Networking}}},
  doi       = {10.1145/3241539.3241548},
  isbn      = {978-1-4503-5903-0},
  langid    = {english},
  month     = oct,
  pages     = {289--304},
  publisher = {Association for Computing Machinery},
  series    = {{{MobiCom}} '18},
  title     = {Towards Environment Independent Device Free Human Activity Recognition},
  urldate   = {2026-03-28},
  year      = 2018
}

@inproceedings{bahadoriReWiSReliableWifi2022,
  author     = {Bahadori, Niloofar and Ashdown, Jonathan and Restuccia, Francesco},
  booktitle  = {2022 {{IEEE}} 23rd {{International Symposium}} on a {{World}} of {{Wireless}}, {{Mobile}} and {{Multimedia Networks}} ({{WoWMoM}})},
  doi        = {10.1109/WoWMoM54355.2022.00027},
  langid     = {english},
  month      = jun,
  pages      = {50--59},
  shorttitle = {{{ReWiS}}},
  title      = {{{ReWiS}}: Reliable Wi-Fi Sensing through Few-Shot Multi-Antenna Multi-Receiver {{CSI}} Learning},
  urldate    = {2026-02-19},
  year       = 2022
}

@article{yinFewSenseScalableCrossdomain2024,
  author  = {Yin, Guolin and Zhang, Junqing and Shen, Guanxiong and Chen, Yingying},
  doi     = {10.1109/TMC.2022.3221902},
  issn    = {1558-0660},
  journal = {IEEE Transactions on Mobile Computing},
  langid  = {english},
  month   = jan,
  number  = {1},
  pages   = {453--468},
  title   = {{{FewSense}}, towards a Scalable and Cross-Domain Wi-Fi Sensing System Using Few-Shot Learning},
  urldate = {2026-03-28},
  volume  = {23},
  year    = 2024
}

@article{shiEnvironmentrobustDevicefreeHuman2022,
  author  = {Shi, Zhenguo and Zhang, J. Andrew and Xu, Richard Yida and Cheng, Qingqing},
  doi     = {10.1109/TMC.2020.3012433},
  issn    = {1558-0660},
  journal = {IEEE Transactions on Mobile Computing},
  langid  = {english},
  month   = feb,
  number  = {2},
  pages   = {541--554},
  title   = {Environment-Robust Device-Free Human Activity Recognition with Channel-State-Information Enhancement and One-Shot Learning},
  urldate = {2026-03-28},
  volume  = {21},
  year    = 2022
}

@article{miaoWifiSensingTechniques2025,
  author     = {Miao, Fucheng and Huang, Youxiang and Lu, Zhiyi and Ohtsuki, Tomoaki and Gui, Guan and Sari, Hikmet},
  doi        = {10.1145/3705893},
  issn       = {0360-0300},
  journal    = {ACM Comput. Surv.},
  langid     = {english},
  month      = jan,
  number     = {5},
  pages      = {107:1--107:30},
  shorttitle = {Wi-Fi Sensing Techniques for Human Activity Recognition},
  title      = {Wi-Fi Sensing Techniques for Human Activity Recognition: Brief Survey, Potential Challenges, and Research Directions},
  urldate    = {2026-03-28},
  volume     = {57},
  year       = 2025
}

@article{meneghelloSHARPEnvironmentPerson2023,
  author     = {Meneghello, Francesca and Garlisi, Domenico and Fabbro, Nicol{\`o} Dal and Tinnirello, Ilenia and Rossi, Michele},
  doi        = {10.1109/TMC.2022.3185681},
  issn       = {1558-0660},
  journal    = {IEEE Transactions on Mobile Computing},
  langid     = {english},
  month      = oct,
  number     = {10},
  pages      = {6161--6175},
  shorttitle = {Sharp},
  title      = {{{SHARP}}: Environment and Person Independent Activity Recognition with Commodity {{IEEE}} 802.11 Access Points},
  urldate    = {2026-02-19},
  volume     = {22},
  year       = 2023
}

@article{liTwostreamConvolutionAugmented2021,
  author    = {Li, Bing and Cui, Wei and Wang, Wei and Zhang, Le and Chen, Zhenghua and Wu, Min},
  copyright = {Copyright (c) 2021 Association for the Advancement of Artificial Intelligence},
  doi       = {10.1609/aaai.v35i1.16103},
  issn      = {2374-3468},
  journal   = {Proceedings of the AAAI Conference on Artificial Intelligence},
  langid    = {english},
  month     = may,
  number    = {1},
  pages     = {286--293},
  title     = {Two-Stream Convolution Augmented Transformer for Human Activity Recognition},
  urldate   = {2026-03-28},
  volume    = {35},
  year      = 2021
}

@inproceedings{ishibashiStoppingCriterionBayesian2023,
  author    = {Ishibashi, Hideaki and Karasuyama, Masayuki and Takeuchi, Ichiro and Hino, Hideitsu},
  booktitle = {Proceedings of {{The}} 26th {{International Conference}} on {{Artificial Intelligence}} and {{Statistics}}},
  issn      = {2640-3498},
  langid    = {english},
  month     = apr,
  pages     = {6463--6497},
  publisher = {PMLR},
  title     = {A Stopping Criterion for Bayesian Optimization by the Gap of Expected Minimum Simple Regrets},
  urldate   = {2026-03-17},
  year      = 2023
}

@misc{akibaOptunaNextgenerationHyperparameter2019,
  archiveprefix = {arXiv},
  author        = {Akiba, Takuya and Sano, Shotaro and Yanase, Toshihiko and Ohta, Takeru and Koyama, Masanori},
  doi           = {10.48550/arXiv.1907.10902},
  eprint        = {1907.10902},
  month         = jul,
  number        = {arXiv:1907.10902},
  primaryclass  = {cs},
  publisher     = {arXiv},
  shorttitle    = {Optuna},
  title         = {Optuna: {{A Next-generation Hyperparameter Optimization Framework}}},
  urldate       = {2026-03-30},
  year          = 2019
}

@inproceedings{bergstraAlgorithmsHyperparameterOptimization2011,
  author    = {Bergstra, James and Bardenet, R{\'e}mi and Bengio, Yoshua and K{\'e}gl, Bal{\'a}zs},
  booktitle = {Advances in {{Neural Information Processing Systems}}},
  langid    = {english},
  pages     = {2546--2554},
  publisher = {Curran Associates, Inc.},
  title     = {Algorithms for Hyper-Parameter Optimization},
  urldate   = {2026-04-28},
  volume    = {24},
  year      = 2011
}

@article{vandermaatenVisualizingDataUsing2008,
  author  = {{Van der Maaten}, Laurens and Hinton, Geoffrey},
  issn    = {1532-4435},
  journal = {Journal of machine learning research},
  langid  = {english},
  number  = {11},
  title   = {Visualizing Data Using T-{{SNE}}},
  volume  = {9},
  year    = 2008
}

@misc{cottiCausallyInterpretableWiFi2026,
  archiveprefix = {arXiv},
  author        = {Cotti, Luca and Lavazza, Luca and Cominelli, Marco and Han, Liying and Dong, Gaofeng and Gringoli, Francesco and Srivastava, Mani B. and Bihl, Trevor and Blasch, Erik P. and Brigham, Daniel O. and Combs, Kara and Kaplan, Lance M. and Cerutti, Federico},
  doi           = {10.48550/arXiv.2604.22979},
  eprint        = {2604.22979},
  month         = apr,
  number        = {arXiv:2604.22979},
  primaryclass  = {cs},
  publisher     = {arXiv},
  title         = {Towards Causally Interpretable {Wi-Fi} {CSI}-Based Human Activity Recognition with Discrete Latent Compression and {LTL} Rule Extraction},
  urldate       = {2026-05-07},
  year          = 2026
}

@misc{sensoyEvidentialDeepLearning2018,
  archiveprefix = {arXiv},
  author        = {Sensoy, Murat and Kaplan, Lance and Kandemir, Melih},
  doi           = {10.48550/arXiv.1806.01768},
  eprint        = {1806.01768},
  month         = oct,
  number        = {arXiv:1806.01768},
  primaryclass  = {cs},
  publisher     = {arXiv},
  title         = {Evidential Deep Learning to Quantify Classification Uncertainty},
  urldate       = {2026-04-19},
  year          = 2018
}
